\PassOptionsToPackage{table,dvipsnames,svgnames,x11names}{xcolor}
\documentclass[10pt,twocolumn,letterpaper]{article}

\usepackage{cvpr}   
\usepackage[accsupp]{axessibility} %

\usepackage[dvipsnames]{xcolor}

\usepackage{adjustbox}

\usepackage{amssymb}%
\usepackage{pifont}%
\newcommand{\cmark}{\ding{51}}%
\newcommand{\xmark}{\ding{55}}%

\definecolor{cvprblue}{rgb}{0.21,0.49,0.74}
\usepackage[pagebackref,breaklinks,colorlinks,citecolor=cvprblue]{hyperref}
\usepackage{adjustbox} %
\usepackage{url}            %
\usepackage{booktabs}       %
\usepackage{amsfonts}       %
\usepackage{nicefrac}       %
\usepackage{microtype}      %
\usepackage{multirow}
\usepackage{multicol}
\usepackage{dsfont}
\usepackage{wrapfig,lipsum}
\usepackage{breakurl}
\usepackage{adjustbox}
\usepackage{blindtext}
\usepackage{float}
\usepackage{tocloft}
\usepackage{amsmath}
\usepackage[normalem]{ulem}
\usepackage[table,dvipsnames,svgnames,x11names]{xcolor}
\newcommand{\ttt}[1]{\texttt{#1}}

\newcommand\blfootnote[1]{
    \begingroup
    \renewcommand\thefootnote{}\footnote{#1}
    \addtocounter{footnote}{-1}
    \endgroup
}
\newcommand\Tstrut{\rule{0pt}
{2.6ex}}         %
\newcommand\Bstrut{\rule[-0.9ex]{0pt}{0pt}}   %
\usepackage{tablefootnote}
\usepackage{soul}
\definecolor{Gray}{gray}{0.95}
\definecolor{gray_1}{HTML}{e7edee}
\definecolor{gray_2}{HTML}{cfd4d5}
\definecolor{gray_3}{HTML}{b7bbbc}
\definecolor{gray_4}{HTML}{a2a5a6}
\definecolor{b_1}{HTML}{d2e2df}
\definecolor{b_2}{HTML}{e1ebe9}
\definecolor{b_3}{HTML}{e9f2f1}
\definecolor{b_4}{HTML}{f3faf9}
\def\paperID{ 7041} %
\def\confName{CVPR}
\def\confYear{2024}
\usepackage[toc,page,header]{appendix}
\usepackage{minitoc}

\title{DeiT-LT: Distillation Strikes Back for Vision Transformer Training on Long-Tailed Datasets}

\author{%
     Harsh Rangwani\textsuperscript{*1} \quad Pradipto Mondal\textsuperscript{*1,2} \quad Mayank Mishra\textsuperscript{*1} \\ \quad Ashish Ramayee Asokan\textsuperscript{1}  \quad R. Venkatesh Babu\textsuperscript{1} \\ \\
     \textsuperscript{1}Indian Institute of Science, Bangalore \quad
      \textsuperscript{2} Indian Institute of Technology, Kharagpur
 }

\begin{document}
\doparttoc %
\faketableofcontents %

\maketitle

\begin{abstract}
Vision Transformer (ViT) has emerged as a prominent architecture for various computer vision tasks. In ViT, we divide the input image into patch tokens and process them through a stack of self-attention blocks. However, unlike Convolutional Neural Network (CNN), ViT’s simple architecture has no informative inductive bias (e.g., locality, etc.). Due to this, ViT requires a large amount of data for pre-training. Various data-efficient approaches (DeiT) have been proposed to train ViT on balanced datasets effectively. However, limited literature discusses the use of ViT for datasets with long-tailed imbalances. In this work, we introduce DeiT-LT to tackle the problem of training ViTs from scratch on long-tailed datasets. In DeiT-LT, we introduce an efficient and effective way of distillation from CNN via distillation \texttt{DIST} token by using out-of-distribution images and re-weighting the distillation loss to enhance focus on tail classes. This leads to the learning of local CNN-like features in early ViT blocks, improving generalization for tail classes.
Further, to mitigate overfitting, we propose distilling from a flat CNN teacher, which leads to learning low-rank generalizable features for \texttt{DIST} tokens across all ViT blocks.  With the proposed DeiT-LT scheme, the distillation \texttt{DIST} token becomes an expert on the tail classes, and the classifier \texttt{CLS} token becomes an expert on the head classes. The experts help to effectively learn features corresponding to both the majority and minority classes using a distinct set of tokens within the same ViT architecture. We show the effectiveness of DeiT-LT for training ViT from scratch on datasets ranging from small-scale CIFAR-10 LT to large-scale iNaturalist-2018.
Project Page: \href{https://rangwani-harsh.github.io/DeiT-LT}{https://rangwani-harsh.github.io/DeiT-LT}.
\end{abstract}
\vspace{-1mm}
   
\section{Introduction}
\label{sec:intro}

Visual Recognition has seen unprecedented success with the advent of deep neural networks trained on large datasets~\cite{deng2009imagenet}. Consequently, efforts are being made to collect large datasets through crowd-sourcing to train deep neural networks for various applications across domains. As a result of crowd-sourcing, these datasets often exhibit long-tailed data distributions due to inherent natural statistics~\cite{van2018inaturalist, gupta2019lvis}, i.e., a large number of images belong to a small portion of (\textit{majority}) classes, whereas other (\textit{minority}) classes contain few image samples each. A lot of recent works~\cite{li2022nested, cao2019learning, cui2019class, zhou2020bbn, menon2020long} focus on training deep neural networks for recognition on such long-tailed datasets, such that networks perform reasonably well across all classes, including the minority classes. Loss manipulation-based techniques~\cite{cao2019learning, cui2019class,kinivs} enhance the network's focus toward learning tail classes by enforcing a large margin or increasing the weight for loss for these classes. As these techniques enhance the focus on the tail classes, they often lead to some performance degradation in the head (majority) classes. To mitigate this, State-of-the-Art (SotA) techniques currently train multiple expert networks~\cite{wang2020long, li2022nested} that specialize in different portions of the data distribution. The predictions from these experts are then aggregated to produce the final output, which improves the performance over individual experts. However, all these efforts have been restricted to Convolutional Neural Networks (CNNs), particularly ResNets~\cite{he2016deep}, with little attention to architectures such as Transformers~\cite{dosovitskiy2015discriminative, vaswani2017attention}, MLP-Mixers~\cite{tolstikhin2021mlp} etc.
\blfootnote{\textsuperscript{*} denotes equal contribution. Correspondence to harshr@iisc.ac.in.}
\begin{figure*}[t]
    \centering
    \includegraphics[width=\textwidth]{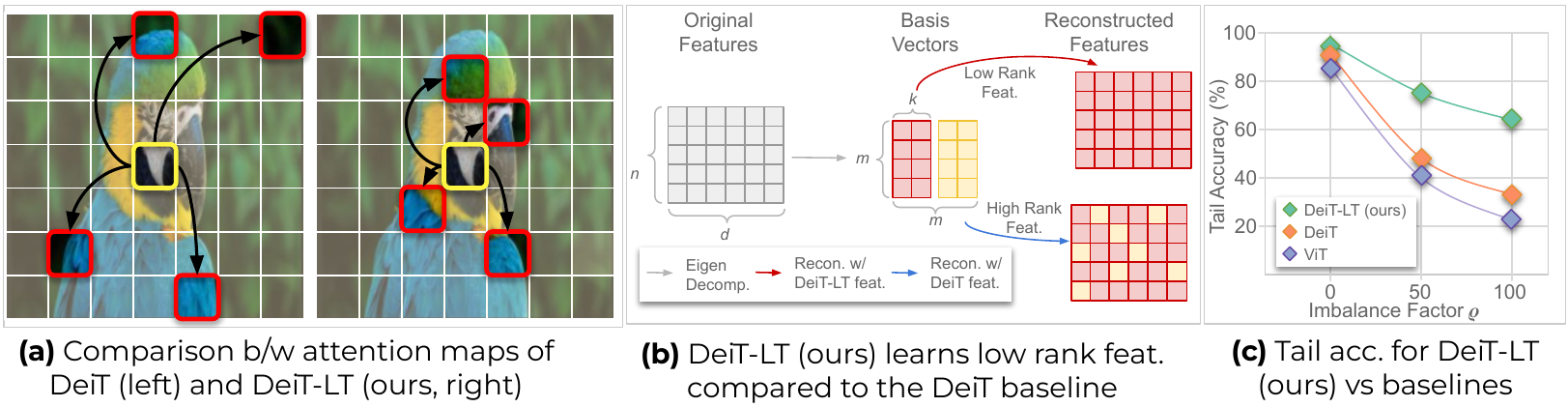}
    \caption{We propose DeiT-LT (Fig.~\ref{fig:overview}, a distillation scheme for Vision Transformer (ViT), tailored towards long-tailed data). In Deit-LT, \textbf{a)} we introduce OOD distillation from CNN, which leads to learning local generalizable features in early blocks. \textbf{b)} we propose to distill from teachers trained via SAM~\cite{foret2020sharpness} which induces low-rank features across blocks in ViT to improve generalization. \textbf{c)} In comparison to other SotA ViT baselines, Deit-LT (ours) demonstrates significantly improved performance for minority classes, with increasing imbalance.}
    \label{fig:teaser-fig}
    \vspace{-1em}
\end{figure*}

Recently, the transformer architecture adapted for computer vision, named as Vision Transformer (ViT)~\cite{dosovitskiyimage}, has gained popularity due to its scalability and impressive performance on various computer vision tasks~\cite{carion2020end, strudel2021segmenter}. One caveat behind its impressive performance is the requirement for pre-training on large datasets~\cite{dosovitskiy2015discriminative}. The data-efficient transformers  (DeiT)~\cite{touvron2021deit} aimed to reduce this requirement for pre-training by distilling information from a pre-trained CNN. Subsequent efforts have further improved the data and compute efficiency ~\cite{touvron2022deit, Touvron2022ThreeTE} of ViTs. However, all these improvements have been primarily based on increasing performance on the balanced ImageNet dataset. We find that these improvements are still insufficient for robust performance on long-tailed datasets (Fig.~\ref{fig:teaser-fig}\textcolor{red}{c}).

In this work, we aim to investigate and improve the \emph{training of Vision Transformers from scratch without the need for large-scale pre-training} on diverse long-tailed datasets, varying in image size and resolution. Recent works show improved performance for ViTs on long-tailed recognition tasks, but they often need expensive pre-training on large-scale datasets~\cite{chen2022reltransformer, long2022retrieval}. The requirement of pre-training is computationally expensive and restricts their application to specialized domains such as medicine, satellite, speech, etc. Furthermore, the large-scale pre-trained datasets often contain biases that might be inadvertently induced with their usage~\cite{agarwal2021evaluating, ousidhoum2021probing, wang2023overwriting}. To mitigate these shortcomings, we introduce \textit{\textbf{Data-efficient Image Transformers for Long-Tailed Data (DeiT-LT)}} - a scheme for training ViTs from scratch on small and large-scale long tailed datasets. DeiT-LT is based on the following important design principles:
\begin{itemize}
    \item DeiT-LT involves distilling knowledge from low-resolution teacher networks using out-of-distribution (OOD) images generated through strong augmentations. Notably, this method proves effective even if the CNN teacher wasn't originally trained on such augmentations. The outcome is the successful induction of CNN-like feature locality in the ViT student network, ultimately enhancing generalization performance, particularly for minority (tail) classes (Fig.~\ref{fig:teaser-fig}\textcolor{red}{a},~\ref{fig:similarity_plot} and Sec.~\ref{sebsec:out-of-dis-dist}).
    
    \item Further, to improve the generality of features, we 
    propose to distill knowledge via flat CNN teachers trained through Sharpness Aware Minimization (SAM)~\cite{foret2020sharpness}. This results in low-rank generalizable features for long-tailed setup across all ViT blocks  (Fig.~\ref{fig:teaser-fig}\textcolor{red}{b} and Sec.~\ref{subsec:low-rank-sam}).
    
    \item In DeiT~\cite{touvron2021deit}, the classification and distillation tokens produce similar predictions. However, in proposed DeiT-LT, we ensure their divergence such that the classification token becomes an expert on the majority classes. Whearas, the distillation token learns local low-rank features, becoming an expert on the minority. Hence, DeiT-LT can focus on both the majority and minority effectively, which is not possible with vanilla DeiT training (Fig.~\ref{fig:attention-vis} and Sec.~\ref{sebsec:out-of-dis-dist}).
\end{itemize} 
We demonstrate the effectiveness of DeiT-LT across diverse small-scale (CIFAR-10 LT, CIFAR-100 LT) as well as large-scale datasets (ImageNet-LT, iNaturalist-2018). We find that DeiT-LT effectively improves over the teacher CNN across all datasets and achieves performances superior to SotA CNN-based methods without requiring any pre-training.

\section{Background}
\vspace{-1mm}
\label{sec:related_works}
\begin{figure*}[!t]
    \centering
    \includegraphics[width=\textwidth]{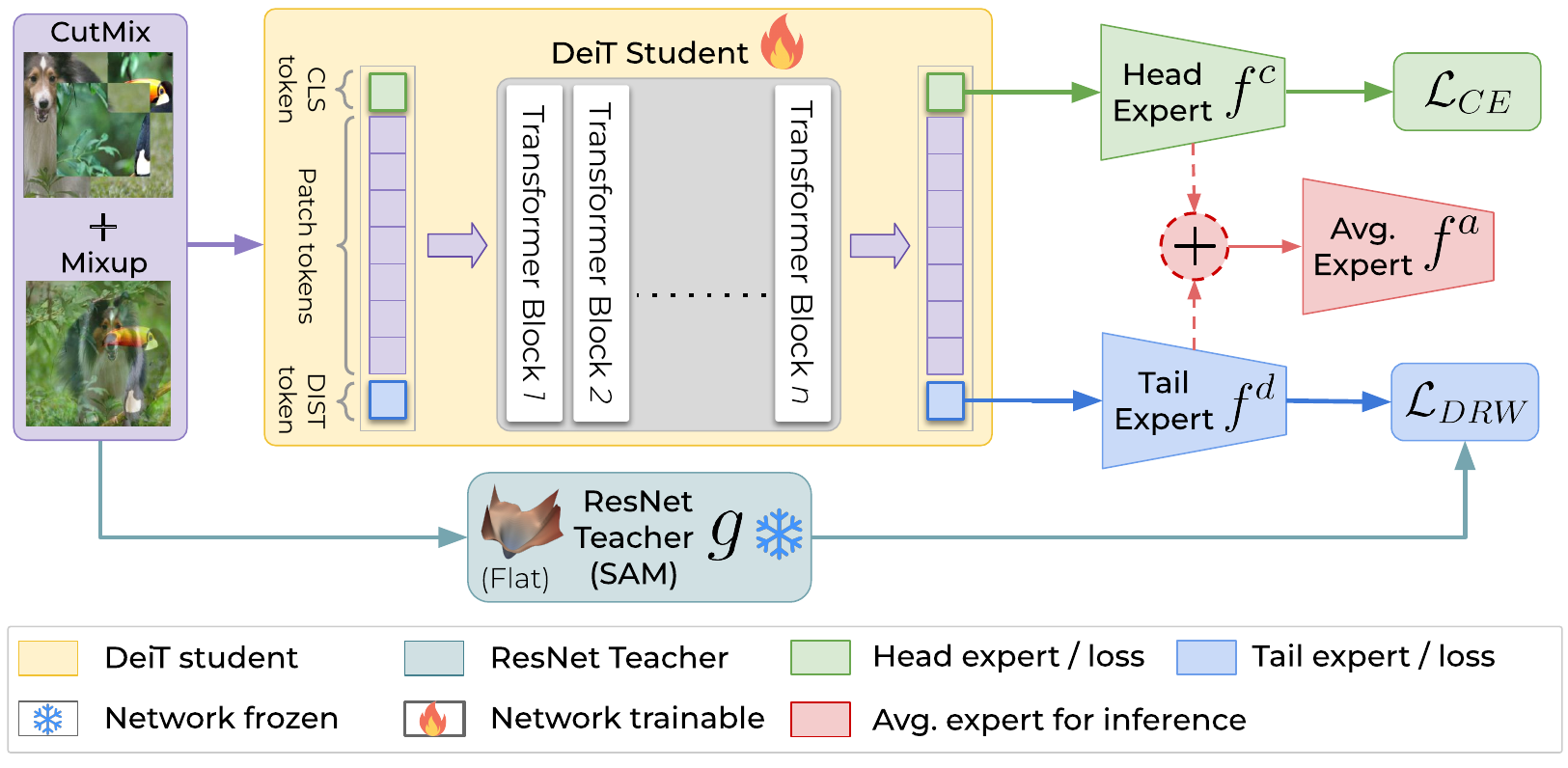}
    \caption{ Overview of DeiT-LT. 
 The Head Expert classifier trains using CE loss against ground truth, whereas the Tail Expert classifier trains using DRW loss against hard-distillation targets from the flat ResNet teacher trained via SAM~\cite{foret2020sharpness}. The distillation is performed using out-of-distribution images created using strong augmentations and Mixup.
 }
    \label{fig:overview}
    \vspace{-1.5 em}
\end{figure*}
\vspace{1mm} \noindent \textbf{Long-Tailed Learning.} With the increased scale of deep learning, large crowd-sourced long-tailed datasets have become common. A plethora of techniques are developed to learn machine learning models using such datasets, where the objective is improved performance, particularly on tail classes. The methods can be broadly divided into three categories: a) loss re-weighting b) decoupled classifier and representations and c) expert-based classifier training. In addition, there are some techniques based on the synthetic generation for long-tailed recognition~\cite{kim2020m2m, rangwani2021class, rangwani2023noisytwins, hrangwani2022gsr}, which are orthogonal to this study. The loss re-weighting-based techniques include margin based techniques like LDAM~\cite{cao2019learning}, and Logit-Adj~\cite{menon2020long}, which enforce a higher margin for tail classes. The other set (eg. CB-Loss~\cite{cui2019class}, VS-Loss~\cite{kinivs} etc.) introduce re-weighting factors in cross entropy loss based on the training set label distribution. The other set of techniques propose to decouple the learning of representations with classifier learning, as it's observed that margin based losses lead to sub-optimal representations~\cite{kang2019decoupling}. The classifier is then learned using Learnable Weight Scaling (LWS), $\tau$-normalization, which improves performance on the tail classes~\cite{kang2019decoupling}. Further, after this follow-up works~\cite{ye2020identifying, wang2021contrastive} like MiSLAS~\cite{zhou2017places} proposed Mixup~\cite{zhang2018mixup} based improved representation learning and LADE~\cite{hong2021disentangling} proposes improved classifier training by adapting to target label distribution. Further, contrastive methods, including PaCo~\cite{cui2021parametric} and BCL~\cite{ren2020balanced}, have demonstrated improved performance with contrastive learning. However, all these methods lead to performance degradation on head classes to improve performance on tail classes.  To mitigate this degradation, the techniques (like RIDE~\cite{wang2020long} etc.) learn different experts on different parts of the data distribution. These experts are learned in a way that makes them diverse in their predictions and can be combined efficiently to obtain improved predictions. However, these methods require additional computation to combine experts at the inference time. In our work, we can efficiently learn experts on majority and minority using a single ViT backbone, the predictions of which we average to prevent any additional inference overhead at the deployment time.

\vspace{1mm} \noindent \textbf{Vision Transformer.} In recent literature, Vision Transformers~\cite{dosovitskiyimage} have emerged as strong competitors for ResNets as they are easier to scale and lead to improved generalization. DeiT~\cite{touvron2021deit} developed a data-efficient way to train these models by distilling through Convolutional Neural Networks. However, despite being data efficient, these models still produce sub-optimal performance on long-tailed data. RAC~\cite{long2022retrieval} utilizes pre-trained transformer for data-efficiency on long-tailed data. However, these pre-trained models are often domain specific and do not generalize well to other domains like medical, synthetic etc. In our work, we train Vision Transformers from scratch, even for small datasets like CIFAR-10 LT, CIFAR-100 LT, which makes them free from biases due to pre-training on large datasets~\cite{wang2023overwriting}.%

\vspace{1mm} \noindent \textbf{Data Efficient Vision Transformers (DeiT).}
The Vision Transformer (ViT) architecture~\cite{dosovitskiyimage} consists of transformer architecture stacked with Multi-Headed Self-Attention blocks~\cite{vaswani2017attention}. To provide input to the Vision Transformer architecture, we first convert the image into patches. These image patches are passed through a linear layer to convert them into tokens that are then passed to the attention blocks. The attention blocks learn the relationship between these tokens for performing a given task. In addition to this, the ViT architecture also contains one classifier (\texttt{CLS}) token that represents the features to be used for classification. In the Data Efficient Transformer (DeiT)~\cite{touvron2021deit}, there is an additional distillation (\texttt{DIST}) token in the ViT backbone that learns via distillation from the teacher CNN. For the classification head and the distillation head,  $\mathcal{L}_{CE}$ is used for training (Fig.~\ref{fig:overview}). The final loss function for the network is:
\begin{equation}
    \mathcal{L}= \mathcal{L}_{CE}(f^{c}(x), y) + \mathcal{L}_{CE} (f^{d}(x), y_{t}), y_t = \arg \max_{i} g(x)_{i}
\end{equation}
Here $f^{c}(x)$ is output from the classifier of student $\texttt{CLS}$ token, $f^{d}(x)$ is output from the classifier of student $\texttt{DIST}$ token, $g(x)$ denotes the output of the teacher CNN network, $y \in [K]$ is the ground truth, $y_t$ is the label produced by the teacher corresponding to the sample $x$, and $N_i$ is the number of samples in class $i$. At the time of inference in DeiT, we obtain logit outputs from the two heads $f^{d}(x)$ and $f^{c}(x)$, and average them to produce the final prediction.

\section{DeiT-LT (DeiT for Long-Tailed Data)}

\begin{table}
\centering
\setlength{\tabcolsep}{7pt}
\caption{\textbf{Effect of augmentations:} Comparison of teacher (\textit{Tch}) and student (\textit{Stu}) accuracy (\%) and training time (in hours) on CIFAR-10 LT ($\rho$ = 100) using various augmentation strategies with mixup (\cmark) and without mixup (\xmark). Despite low teacher training accuracy on the out-of-distribution images, the student (Stu.) performs better on the validation set.}
\label{tab:augs}
\vspace{-2mm}
\resizebox{\linewidth}{!}{%
\begin{tabular}{@{}c|cc|cc|c@{}}
\toprule
\textbf{\begin{tabular}[c]{@{}c@{}}Tch\\ Model\end{tabular}} &
  \textbf{\begin{tabular}[c]{@{}c@{}}Stu\\ Augs.\end{tabular}} &
  \textbf{\begin{tabular}[c]{@{}c@{}}Tch\\ Augs.\end{tabular}} &
  \textbf{\begin{tabular}[c]{@{}c@{}}Tch\\ Acc.\end{tabular}} &
  \textbf{\begin{tabular}[c]{@{}c@{}}Stu\\ Acc.\end{tabular}} &
  \textbf{\begin{tabular}[c]{@{}c@{}}Train\\ Time\end{tabular}} \\ \midrule
\begin{tabular}[c]{@{}c@{}}RegNetY\\ 16GF\end{tabular} & Strong (\cmark) & Strong (\cmark) & 79.1 & 70.2 & 33.3 \\ \midrule
\multirow{3}{*}{ResNet-32}                             & Strong (\xmark) & Weak (\xmark)   & 97.2 & 54.2 & 17.8 \\
                                                       & Strong (\xmark) & Strong (\xmark) & 71.9 & 69.6 & 17.8 \\
                                                       & Strong (\cmark) & Strong (\cmark) & 56.6 & 79.4 & 19.0 \\ \bottomrule
\end{tabular}}
\vspace{-4mm}
\end{table}    

\label{sec:deit_lt}
In this section, we introduce DeiT-LT - the Data-efficient Image Transformer that is specialized to be effective for Long-Tailed data. We start with a DeiT transformer-based architecture which, in addition to the classification (\ttt{CLS}) token, also contains a distillation (\ttt{DIST}) token (Fig.~\ref{fig:overview}) that learns via distillation from a CNN.
The DeiT-LT introduces three particular design components, which are: \textbf{a)} the effective distillation via out-of-distribution (OOD) images, which induces local features and leads to the creation of experts,  \textbf{b)} training Tail Expert classifier using DRW loss and \textbf{c) }learning of low-rank generalizable features from flat teachers via distillation. In the following sections, we analyze our design choices in detail. We analyze CIFAR-10 LT using LDAM+DRW+SAM ResNet-32~\cite{rangwani_escapingsaddle} CNN teacher, to justify the rationale behind each design component.

\subsection{ Distillation via Out of Distribution Images}
\label{sebsec:out-of-dis-dist}
We now focus on how to distill knowledge from a CNN architecture to a ViT effectively. In the original DeiT work \cite{touvron2021deit}, the authors first train a large CNN, specifically RegNetY~\cite{radosavovic2020designing}, with strong augmentations ($\mathcal{A}$) as used by a ViT for distillation. However, this incurs the additional expense of training a large CNN for subsequent training of the ViT through distillation. In contrast, we propose to train a small teacher CNN (ResNet-32) with the usual weak augmentations, but during distillation, we pass strongly augmented images to obtain predictions to be distilled. 

\begin{figure}[!t]
\centering
    \includegraphics[width=0.9\linewidth]{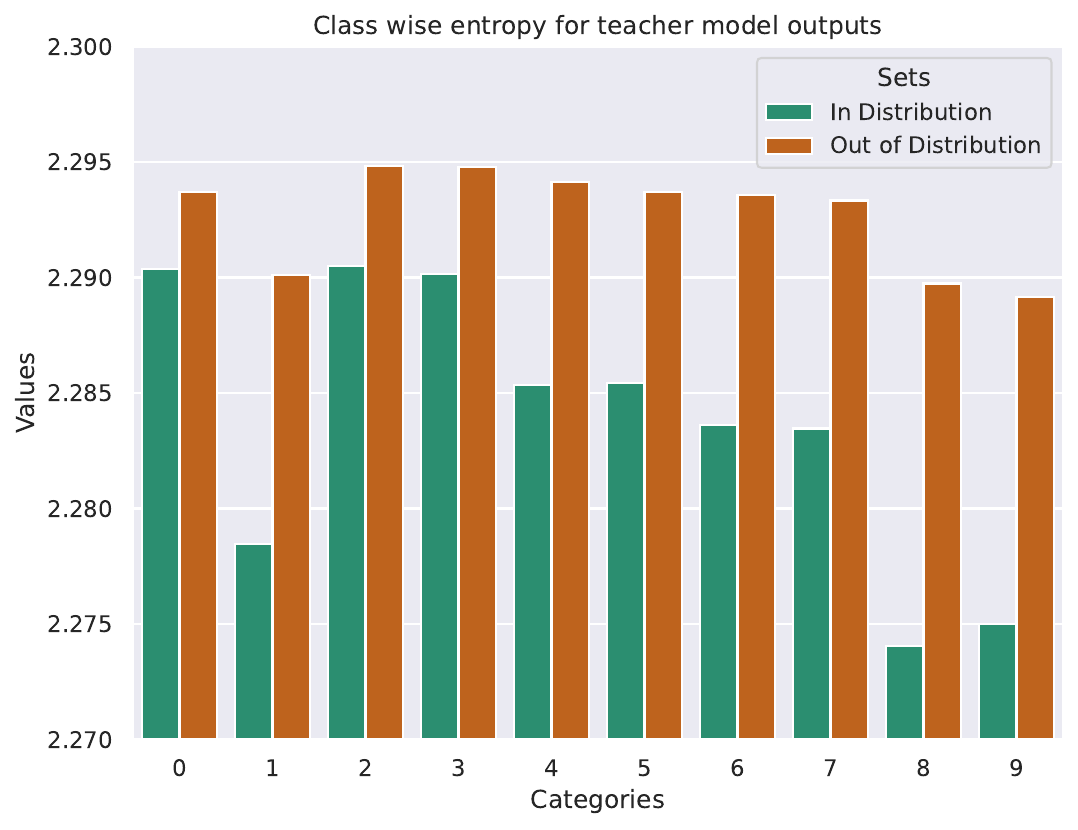}
    \vspace{-2mm}
    \caption{\textbf{Entropy of teacher outputs:} Comparison of the entropy of in-distribution samples and out-of-distribution samples with the ResNet-32 teacher on CIFAR-10 LT. We observe a higher accuracy in Table-\ref{tab:augs} corresponding to out-of-distribution samples.}
    \label{fig:entropy}
    \vspace{-2mm}
\end{figure}

These strongly augmented images are \emph{out-of-distribution (OOD)} images for the ResNet-32 CNN as the model's accuracy on these training images is low, as seen in Table \ref{tab:augs}. However, despite the low accuracy, the strong augmentations lead to effective distillation in comparison to the weak augmentations on which the original ResNet was trained (Table \ref{tab:augs}). This works because the ViT student learns to mimic the incorrect predictions of the CNN teacher on the out-of-distribution images, which in turn enables the student to learn the inductive biases of the teacher.

\begin{equation}
    f^{d} (X) \approx g (X) , X \sim A(x)
\end{equation}
Further, we find that creating additional out-of-distribution samples by mixing up images from two classes~\cite{yun2019cutmix, zhang2018mixup} improves the distillation performance. This can also be seen

\noindent
from the entropy of predictions on teacher, which are high (\ie more informative) for OOD samples (Fig.~\ref{fig:entropy}). \textit{In general, we find that increasing diverse amount of out-of-distribution~\cite{nayak2021effectiveness} data while distillation helps improve performance and leads to effective distillation from the CNN.} Details regarding the augmentations are in Suppl. Sec. \ref{suppl:training_aug}.

\begin{figure*}[!t]
     \centering
     \begin{subfigure}[b]{0.32\textwidth}
         \centering
         \includegraphics[width=\textwidth]{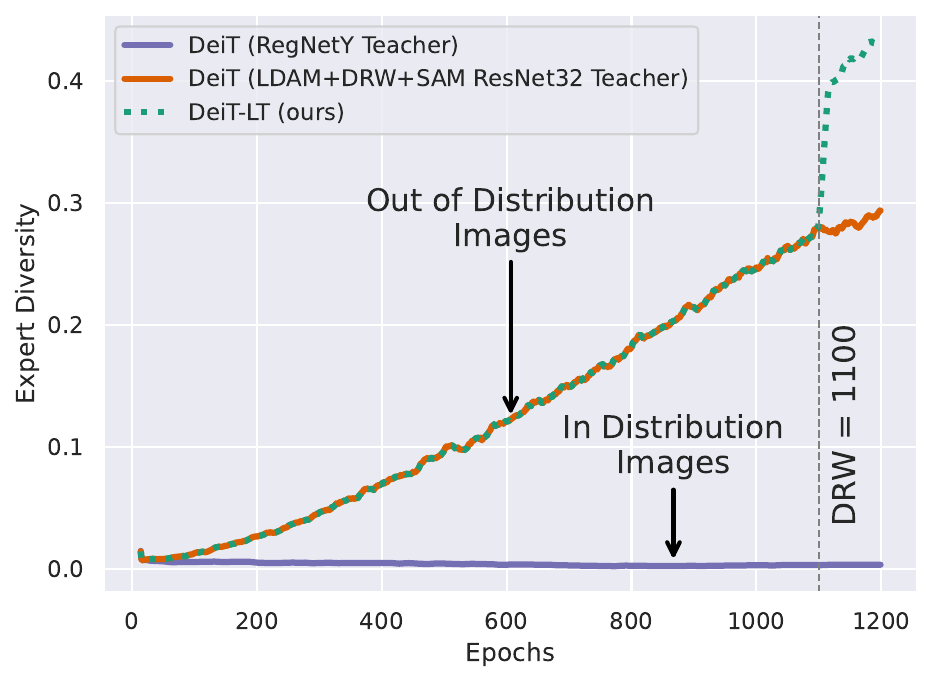}
         \caption{Diversity for \ttt{CLS} and \ttt{DIST} experts}
         \label{fig:similarity_plot}
     \end{subfigure}
     \begin{subfigure}[b]{0.29\textwidth}
         \centering
         \includegraphics[width=\textwidth]{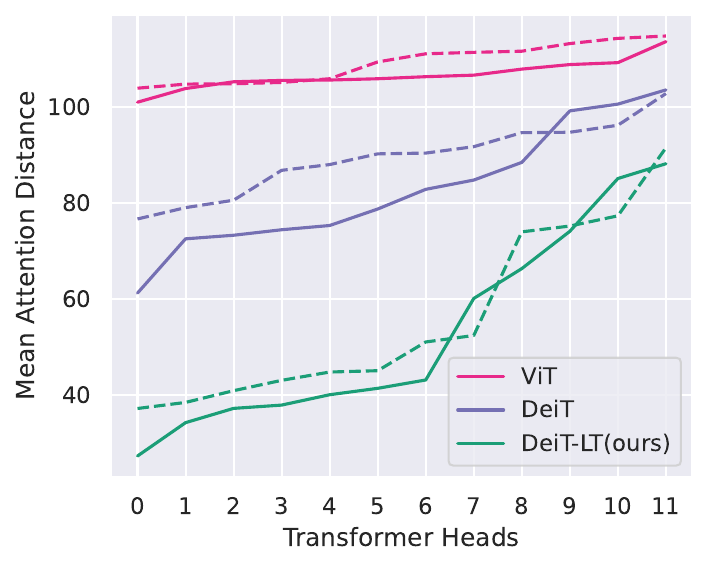}
         \caption{Locality of Attention Heads}
         \label{fig:ViT_locality}
     \end{subfigure}
    \begin{subfigure}[b]{0.32\textwidth}
         \centering
         \includegraphics[width=\textwidth]{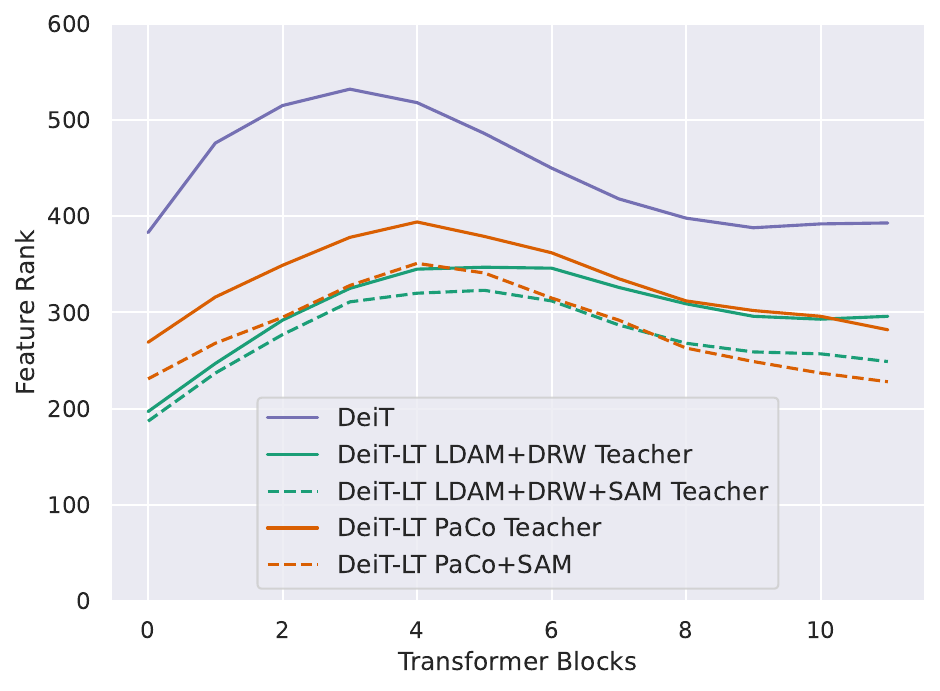}
         \caption{Rank of ViT from Distillation of CNN teachers.}
         \label{fig:ViT_rank}
     \end{subfigure}
     
        \caption{Effect of Distillation in DeiT-LT.  In \textbf{a)} we train DeiT-B with teachers trained on in-distribution images (RegNetY-16GF) and out-of-distribution images (ResNet32). The out-of-distribution distillation leads to diverse experts, which become more diverse with deferred re-weighting on the distillation token (DRW). In \textbf{b)} we plot the \emph{Mean Attention Distance} for the patches across the early self attention block 1 (solid) and  block 2 (dashed) for baselines, where we find that DeiT-LT leads to highly local and generalizable features. In \textbf{c)} we show the rank of features for \ttt{DIST} token, where we demonstrate that students trained with SAM are more low-rank in comparison to baselines}
        \label{fig:dist_sim}
\vspace{-2mm}
\end{figure*}

 Due to distillation via out-of-distribution images, the teacher predictions $y_t$ often differ from the ground truth $y$. Hence, the classification token (\texttt{CLS}) and distillation token (\texttt{DIST}) representations diverge while training. This phenomenon can be observed in Fig. \ref{fig:similarity_plot}, where the cosine distance between the representation of the \texttt{CLS} and \texttt{DIST} tokens increases as the training progresses. This leads to the \texttt{CLS} token being an expert on head classes, while the \texttt{DIST} token specializes in tail class predictions. Our observation debunks the \emph{myth that it is required for the \texttt{CLS} token predictions to be similar to \texttt{DIST}} for effective distillation in transformer, as observed by \citet{touvron2021deit}. 

 \noindent \textbf{Tail Expert with DRW loss.} Further in this stage, we also introduce Deferred Re-Weighting (DRW)~\cite{cao2019learning} for distillation loss, where we weigh the loss for each class using a factor $w_y = {1}/\{1 + (e_y - 1)\mathds{1}_{\mathrm{epoch \geq K}}\}$, where $e_y=\frac{1-\beta^{N_y}}{1-\beta}$ is the effective number of samples in class $y$~\cite{cui2019class}, after $K$ number of epochs~\cite{cao2019learning}. Hence the overall loss is given as:  

\begin{align*}
    \mathcal{L} = \frac{1}{2} \mathcal{L}_{CE}(f^{c}(x), y) + \frac{1}{2}\mathcal{L}_{DRW} (f^{d}(x), y_{t}),\\
     \text{where} ~ \mathcal{L}_{DRW} = -w_{y_t} \; log (f^{d}(x)_{y_t})
\end{align*}
The DRW stage further enhances the focus of the distillation head (\texttt{DIST}) on the tail classes, leading to improved performance. This is also observed in Fig. \ref{fig:similarity_plot}, where the diversity between the two tokens improves after the introduction of the DRW stage. This leads to the creation of diverse \ttt{CLS} and \ttt{DIST} tokens, which are experts on the majority and minority classes respectively.

\noindent \textbf{Induction of Local Features: } To gain insights into the generality and effectiveness of OOD Distillation, we take a closer look at the tail features produced by DeiT-LT. In Fig.~\ref{fig:ViT_locality}, we plot the mean attention distance for each patch across ViT heads~\cite{raghu2021vision} (Details in Suppl. Sec. \ref{suppl:local_connectivity}).  

\vspace{1mm} \noindent \textbf{Insight 1:} DeiT-LT contains heads that attend locally, like CNN, in the neighborhood of the patch in early blocks (1,2). 

\vspace{1mm} \noindent Due to this learning of local generalizable class agnostic features, we observe improved generalization on minority classes (Fig. \ref{fig:teaser-fig}\textcolor{red}{c}). Without the OOD distillation, we find that the vanilla DeiT-III and ViT baselines overfit only on the spurious global features (Fig.~\ref{fig:ViT_locality}) and do not generalize well for tail classes. Hence, this makes OOD distillation in DeiT-LT a well-suitable method for long-tailed scenarios.

\subsection{Low-Rank Features via SAM teachers}
\label{subsec:low-rank-sam}
To further improve the generalizability of the features, particularly for classes with less data, we propose to distill via \emph{teacher CNN models trained via Sharpness Aware Minimization (SAM) objective}~\cite{foret2020sharpness}. Models trained via SAM  converge to flat minima~\cite{rangwani_escapingsaddle} and lead to low-rank features~\cite{andriushchenko2023sharpness}. For analyzing the rank of features for the ViT student in LT case, we calculate rank specifically for the features of tail classes~\cite{andriushchenko2023sharpness}. We detail the procedure of our rank calculation in Suppl. Sec. \ref{suppl:low_rank}.
We confirm our observations across diverse teacher models trained via LDAM and PaCo.
We find the following insight for distillation via \ttt{DIST} token: 

\vspace{1mm} \noindent \textbf{Insight 2.} We observe that distilling into ViT via predictions made using SAM teacher leads to low-rank generalizable (\texttt{DIST}) token features across blocks of ViT (Fig. \ref{fig:ViT_rank}). 

\vspace{1mm}  \noindent {This transfer of a CNN teacher's characteristic (low-rank) to the student, by just distilling via final logits, is a significant novel finding in the context of distillation for ViTs.}

\vspace{1mm} \noindent \textbf{Training Time.} In the original DeiT formulation, the authors~\cite{touvron2022deit} propose training a large CNN RegNetY-16GF at a high resolution (224 $\times$ 224) for distillation to the ViT. We find that competitive performance can be achieved even with training a smaller ResNet-32 CNN (32 $\times$ 32) at a lower resolution, as seen in Table~\ref{tab:augs}. This significantly reduces compute requirement and overall training time by 13 hours, as the ResNet-32 model can be trained quickly (Table~\ref{tab:augs}). Further, we find that with SAM teachers, the student converges much faster than vanilla teacher models, demonstrating the efficacy of SAM teachers for low-rank distillation (Suppl. Sec. \ref{suppl:sam_convergence}).

\section{Experiments}

\subsection{Datasets}

We analyze the performance of our proposed method on four datasets, namely \textbf{CIFAR-10 LT}, \textbf{CIFAR-100 LT}, \textbf{ImageNet-LT}, and \textbf{iNaturalist-2018}. We follow \cite{cao2019learning} to create long-tailed versions of CIFAR~\cite{krizhevsky2009learning} datasets, where the number of samples is exponentially decayed using an imbalance factor $\rho =\frac{\max_i N_{i}}{\min_j N_j}$ (number of samples in the most frequent class by that in the least frequent class). For ImageNet-LT, we create an imbalanced version of the ImageNet~\cite{russakovsky2015imagenet} dataset as described in \cite{liu2019large}. We also report performance on iNaturalist-2018~\cite{van2018inaturalist}, a real-world long-tailed dataset. We divide the classes into three subcategories: \textbf{Head} (\textit{Many}), \textbf{Mid} (\textit{Medium}), and \textbf{Tail} (\textit{Few}) classes. More details regarding the datasets can be found in Suppl. Sec. \ref{suppl:datasets}.
\begin{table}[!t]
    \centering
    
    \caption{Results on CIFAR-10 LT and CIFAR-100 LT datasets with $\rho$=50 and $\rho$=100. We report the \emph{overall} accuracy for available methods. (The teacher used to train the respective student (DeiT-LT) model can be identified by matching superscripts)}
    \resizebox{\linewidth}{!}{%
    \begin{tabular}{l|c|c|c|c}
        \toprule[1pt]
         \multirow{2}{*}{\begin{tabular}{c}\textbf{Method}\end{tabular}} & \multicolumn{2}{c|}{CIFAR-10 LT} & \multicolumn{2}{c}{CIFAR-100 LT} \Tstrut\Bstrut\\
         \cline{2-5}
         & \multicolumn{1}{c|}{$\rho$ = 100} & \multicolumn{1}{c|}{$\rho$ = 50} & \multicolumn{1}{c|}{$\rho$ = 100} & \multicolumn{1}{c}{$\rho$ = 50} \Tstrut\Bstrut\\
         \midrule
         \rowcolor{Gray} \multicolumn{5}{c}{ResNet32 Backbone} \Tstrut\\
         \midrule
         
         CB Focal loss~\cite{cui2019class} & 74.6 & 79.3 & 38.3 & 46.2\\ 
         LDAM+DRW~\cite{cao2019learning} & 77.0 & 79.3 & 42.0 & 45.1 \\ 
         LDAM+DAP~\cite{jamal2020rethinking} & 80.0 & 82.2 & 44.1 & 49.2 \\
         BBN~\cite{zhou2020bbn} & 79.8 & 82.2 & 39.4 & 47.0 \\
         
        CAM~\cite{zhang2021bag} & 80.0 & 83.6  & 47.8 & 51.7 \\
        Log. Adj.~\cite{menon2020long} & 77.7 & -  & 43.9 & - \\
        RIDE~\cite{wang2020long}  & - & -  & 49.1 & - \\
        MiSLAS~\cite{zhong2021improving} & 82.1 & 85.7  & 47.0 & 52.3 \\
        Hybrid-SC~\cite{wang2021contrastive} & 81.4 & 85.4  & 46.7 & 51.9 \\
        SSD~\cite{li2021self_iccv} & - & -  & 46.0 & 50.5\\
        ACE~\cite{cai2021ace} & 81.4 & 84.9  & 49.6 & 51.9 \\
        GCL~\cite{gcl} & 82.7 & 85.5 & 48.7 & 53.6\\ 
    
        VS~\cite{kinivs} & 78.6 & -  & 41.7 \\
        VS+SAM~\cite{rangwani_escapingsaddle} & 82.4 & - & 46.6 & - \\
\midrule  
        \rowcolor{b_1}$^{1}$\small{L-D-SAM} \cite{rangwani_escapingsaddle} & 81.9 & 84.8 & 45.4  & 49.4 \\
        \rowcolor{b_2}$^{2}$PaCo+SAM\cite{rangwani_escapingsaddle, cui2021parametric} & 86.8  & 88.6  & 52.8& 56.6  \\
        \midrule

        \rowcolor{Gray} \multicolumn{5}{c}{ViT-B Backbone} \\
         \midrule  
         ViT ~\cite{dosovitskiyimage} & 62.6 & 70.1 & 35.0 & 39.0 \\
          ViT (cRT) ~\cite{kang2019decoupling} & 68.9 & 74.5 & 38.9 & 42.2 \\
         DeiT ~\cite{touvron2021deit} & 70.2 & 77.5 & 31.3 & 39.1 \\
         DeiT-III ~\cite{touvron2022deit} & 59.1 & 68.2 & 38.1 & 44.1\\

         \midrule
         \rowcolor{b_1}$^{1}$DeiT-LT(ours) & 84.8 & 87.5 & 52.0 & 54.1 \\
         \rowcolor{b_2}$^{2}$DeiT-LT(ours) & \textbf{87.5} & \textbf{89.8} & \textbf{55.6} & \textbf{60.5} \\

\bottomrule

    \end{tabular}}
    \label{tab:cifar10_cifar100}
    \vspace{-4mm}
    \end{table}

\subsection{Experimental Setup}
We follow the setup mentioned in DeiT~\cite{touvron2021deit} to create the student backbone for our experiments. We use the DeiT-B student backbone architecture for all the datasets.
We train our teacher models using re-weighting based LDAM-DRW-SAM method~\cite{rangwani_escapingsaddle} and the contrastive PaCo+SAM (training PaCo~\cite{cui2021parametric} with SAM~\cite{foret2020sharpness} optimizer), employing ResNet-32 for small scale datasets (CIFAR-10 LT and CIFAR-100 LT) and ResNet-50 for large scale ImageNet-LT, and iNaturalist-2018.
We train the head expert classifier with CE loss $\mathcal{L}_{CE}$ against the ground truth, while the tail expert classifier is trained with the CE+DRW loss $\mathcal{L}_{DRW}$ against the hard-distillation targets from the teacher network. 

\vspace{1mm} \noindent \textbf{Small scale CIFAR-10 LT and CIFAR-100 LT.} These models are trained for 1200 epochs, where DRW training for the Tail Expert Classifier starts from epoch 1100. Except for the DRW training (last 100 epochs), we use Mixup and Cutmix augmentation for the input images. These datasets are trained with a cosine learning rate schedule with a base LR of $5 \times 10^{-4}$ using the AdamW~\cite{loshchilov2017decoupled} optimizer.

\vspace{1mm} \noindent \textbf{Large scale ImageNet-LT and iNaturalist-2018.} These models are trained for 1400 and 1000 epochs, respectively, with the DRW training for the Tail Expert Classifier starting from 1200 and 900 epochs. We use Mixup and Cutmix throughout training. Both datasets follow a cosine learning rate schedule, with a base LR of $5 \times 10^{-4}$. More details on the experimental process can be found in Suppl. Sec~\ref{suppl:experimental}.

\vspace{1mm} \noindent \textbf{Baselines.} We use the popular data-efficient baselines for ViT: \textbf{a) ViT: }The standard Vision Transformer (\textbf{ViT-B})\cite{dosovitskiyimage}   architecture trained with CE Loss against the ground truth. For a fair comparison, we train ViT with the same augmentation strategy used for the DeiT-LT experiments. \textbf{b) DeiT~\cite{touvron2021deit}}: Vanilla DeiT model that uses RegNetY-16GF teacher trained with in-distribution images for distillation. \textbf{c) DeiT-III: } A recent improved version of DeiT (\cite{touvron2022deit}) that focuses on improving the supervised learning of ViT on balanced datasets using three simple augmentations (GrayScale, Solarisation, and Gaussian Blur) and LayerScale \cite{touvron2021going}, also demonstrating the redundancy of distillation in DeiTs. The long-tailed baseline of \textbf{d) ViT (cRT): } a decoupled approach of first training classifier (ViT) and then re-training the classifier for a small number of epochs with class-balanced sampling \cite{kang2019decoupling}. We further attempted training other baselines like LDAM, etc, on ViT. However, we found some optimization difficulties in training ViTs (details in Suppl. Sec. \ref{suppl:additional_baselines}).

We want to convey that we do not compare against baselines~\cite{tian2022vl,long2022retrieval, LiVT, xu2023rethink}, which use pre-training, usually on large datasets, to produce results on even CIFAR datasets (Ref. Suppl. Sec. \ref{suppl:clip}). Our goal is to develop a generic technique for training ViTs across domains and modalities on long-tailed data without requiring any external supervision.

\vspace{-1mm}
\section{Results}
\vspace{-1mm}
In this section, we present results for DeiT-LT across various datasets. We use re-weighting based LDAM+DRW+SAM (referred to as L-D-SAM in Table \ref{tab:cifar10_cifar100},\ref{tab:imgnet},\ref{tab:inat}) and contrastive PaCo+SAM teachers for training DeiT-LT student models.

\begin{table}[!t]
    \centering
    \setlength{\tabcolsep}{7pt}
    \caption{Results on ImageNet-LT. (The teacher used to train respective student (DeiT-LT) can be identified by matching superscripts) }
    
    \resizebox{\linewidth}{!}{%
        \begin{tabular}{l|c|ccc}
            \toprule[1pt]
            \multirow{3}{*}{Method} & \multicolumn{4}{c}{ImageNet-LT} \Bstrut \\
            \cline{2-5}
            & Overall & Head & Mid & Tail \Tstrut\Bstrut\\
            \midrule
            \rowcolor{Gray} \multicolumn{5}{c}{ResNet50 Backbone} \Tstrut\Bstrut\\
            \midrule
            CB Focal loss~\cite{cui2019class} & 33.2 & 39.6 & 32.7 & 16.8 \\ 
            LDAM~\cite{cao2019learning} & 49.8 & 60.4 & 46.9 & 30.7 \\ 
            c-RT~\cite{kang2019decoupling} & 49.6 & 61.8 & 46.2 & 27.3 \\
            $\tau$-Norm~\cite{kang2020exploring} & 49.4 & 59.1 & 46.9 & 30.7 \\
            Log. Adj.~\cite{menon2020long} & 50.1 & 61.1 & 47.5 & 27.6 \\
            RIDE(3 exps)~\cite{wang2020long} & 54.9 & 66.2 & 51.7 & 34.9 \\
            MiSLAS~\cite{zhong2021improving} & 52.7 & 62.9 & 50.7 & 34.3 \\
            Disalign~\cite{zhang2021distribution} & 52.9 & 61.3 & 52.2 & 31.4 \\ 
            TSC~\cite{li2022targeted} & 52.4 & 63.5 & 49.7 & 30.4 \\ 
            GCL~\cite{gcl} & 54.5 & 63.0 & 52.7 & 37.1 \\
            SAFA~\cite{hong2022safa} & 53.1 & 63.8 & 49.9 & 33.4 \\
            BCL~\cite{ren2020balanced} & 57.1 & 67.9 & 54.2 & 36.6 \\
            ImbSAM~\cite{zhou2023imbsam} & 55.3 & 63.2 & 53.7 & 38.3 \\
            CBD$_{ENS}$~\cite{iscen2021cbd}& 55.6&68.5 &52.7 &29.2 \\
            \midrule
            \rowcolor{b_1}$^{1}$L-D-SAM~\cite{rangwani_escapingsaddle} & 53.1 & 62.0 & 52.1 & 32.8 \\
            \rowcolor{b_2}$^{2}$PaCo+SAM~\cite{rangwani_escapingsaddle, cui2021parametric} & 57.5	& 62.1	& 58.8	& 39.3 \\
            \midrule
            \rowcolor{Gray} \multicolumn{5}{c}{ViT-B Backbone} \\
            \midrule
            ViT ~\cite{dosovitskiyimage} & 37.5 & 56.9 & 30.4 & 10.3 \\
            DeiT-III ~\cite{touvron2022deit} & 48.4 & \textbf{70.4} & 40.9 & 12.8 
            
            \\
            \midrule
            \rowcolor{b_1}$^{1}$DeiT-LT(ours) & 55.6 & 65.2 & 54.0 & 37.1 \\
            \rowcolor{b_2}$^{2}$DeiT-LT(ours) & \textbf{59.1} & 66.6 & \textbf{58.3} & \textbf{40.0} \\
            \bottomrule
        \end{tabular}
    }
    \label{tab:imgnet}
    \vspace{-4mm}
\end{table}

\begin{table}[!t]
    \centering
    \setlength{\tabcolsep}{10pt}
    \caption{Results on iNaturalist-2018. (The teacher used to train student (DeiT-LT) can be identified by matching superscripts) }
    
    \resizebox{\linewidth}{!}{%
        \begin{tabular}{l|c|ccc}
            \toprule[1pt]
            \multirow{3}{*}{Method} & \multicolumn{4}{c}{iNaturalist-2018} \Bstrut \\
            \cline{2-5}
            & Overall & Head & Mid & Tail \Tstrut\Bstrut\\
            \midrule
            \rowcolor{Gray} \multicolumn{5}{c}{ResNet50 Backbone} \Tstrut\Bstrut\\
            \midrule
            c-RT~\cite{kang2019decoupling} & 65.2 & 69.0 & 66.0 & 63.2 \\
            $\tau$-Norm~\cite{kang2020exploring} & 65.6 & 65.6 & 65.3 & 65.9  \\
            RIDE(3 exps)~\cite{wang2020long} & 72.2 & 70.2 & 72.2 & 72.7 \\
            MiSLAS~\cite{zhong2021improving} & 71.6 & \textbf{73.2} & 72.4 & 70.4 \\
            Disalign~\cite{zhang2021distribution} & 70.6 & 69.0 & 71.1 & 70.2 \\ 
            TSC~\cite{li2022targeted} & 69.7 & 72.6 & 70.6 & 67.8 \\ 
            GCL~\cite{gcl} & 71.0 & 67.5 & 71.3 & 71.5 \\
            ImbSAM~\cite{zhou2023imbsam}& 71.1 & 68.2 & 72.5 & 72.9 \\
            CBD$_{ENS}$~\cite{iscen2021cbd} & 73.6 & 75.9 & 74.7 & 71.5 \\
            \midrule
            \rowcolor{b_1}$^{1}$L-D-SAM~\cite{rangwani_escapingsaddle} & 70.1 & 64.1 & 70.5 & 71.2 \\
            \rowcolor{b_2}$^{2}$PaCo+SAM~\cite{rangwani_escapingsaddle} & 73.4 & 66.3 & 73.6 & 75.2 \\
            \midrule
            \rowcolor{Gray} \multicolumn{5}{c}{ViT-B Backbone} \\
            \midrule
            ViT ~\cite{dosovitskiyimage} & 54.2 & 64.3 & 53.9 & 52.1 \\
            DeiT-III ~\cite{touvron2022deit} & 61.0 & 72.9 & 62.8 & 55.8 \\
            \midrule
            \rowcolor{b_1}$^{1}$DeiT-LT(ours) & 72.9 & 69.0 & 73.3 & 73.3 \\
            \rowcolor{b_2}$^{2}$DeiT-LT(ours) & \textbf{75.1} & 70.3 & \textbf{75.2} & \textbf{76.2} \\
            \bottomrule
        \end{tabular}
    }
    \label{tab:inat}
    \vspace{-5mm}
\end{table}

\begin{figure}[!t]
    \centering
    \includegraphics[width=\columnwidth]{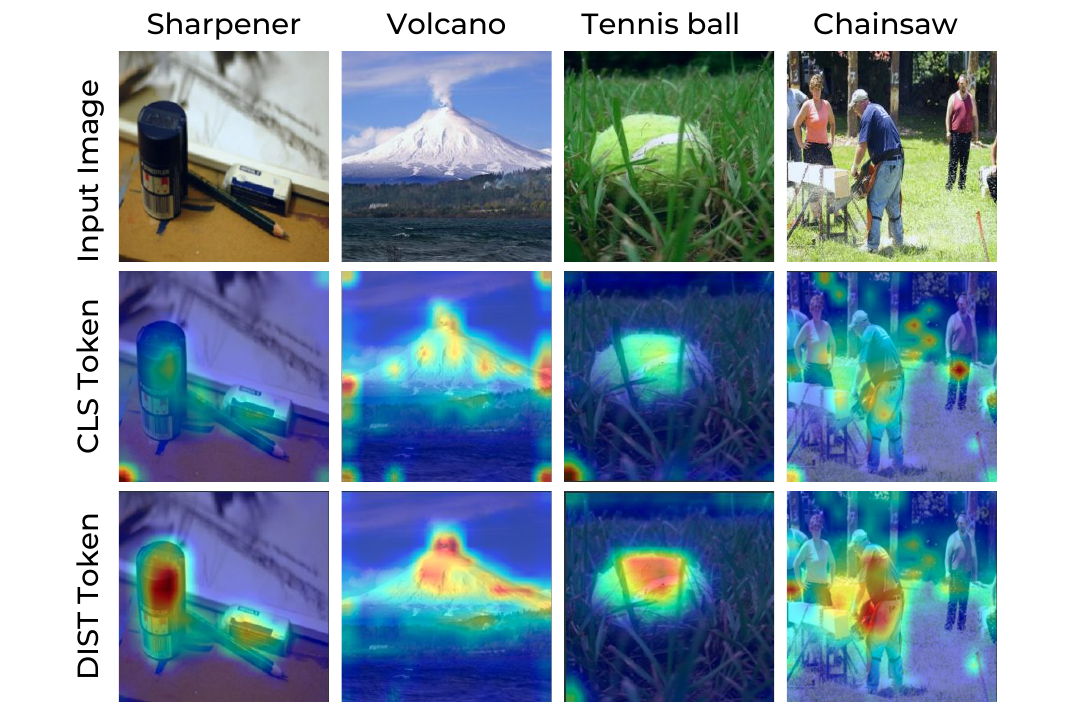}
    \caption{ Visual comparison of the attention maps with respect to the \ttt{CLS} and \ttt{DIST} tokens for \emph{tail} images from the ImageNet-LT dataset. The attention maps are computed by \emph{Attention Rollout} \cite{abnar2020quantifying}.}
    \label{fig:attention-vis}
    \vspace{-5mm}
\end{figure}

\vspace{1mm} \noindent \textbf{Results on Small Scale Datasets.} 
Table \ref{tab:cifar10_cifar100} presents results for the CIFAR-10 LT and CIFAR-100 LT datasets, with varying imbalance factors ($\rho = 100$ and $\rho = 50$). We primarily compare our results to the SotA methods, which train the networks from scratch. The other techniques utilize additional pre-training with extra data~\cite{LiVT, chen2022reltransformer}, making the comparison unfair. Our proposed student network DeiT-LT outperformed the teachers used for their training by an average of 1.9\% and 4.5\% on CIFAR-10 LT and CIFAR-100 LT, respectively. 
This demonstrates the advantage of training the DeiT-LT transformer, which provides additional generalization improvements over the CNN teacher. Further, the DeiT-LT (PaCo+SAM) model significantly improves by 24.9\%  over the ViT baseline (which has the same augmentations as in DeiT-LT) and 28.4\% over the data efficient DeiT-III transformer for CIFAR-10 LT dataset for $\rho = 100$. A similar improvement can also be observed for the CIFAR-100 LT dataset, where DeiT-LT (PaCo+SAM) fares better than ViT baseline and DeiT-III by 20.6\% and 17.5\%, respectively. This shows the effectiveness of the DeiT-LT distillation procedure via CNN teachers. Compared to CNN-based methods, we demonstrate that the transformer-based methods can achieve SotA performance when trained with DeiT-LT distillation procedure, combining both the scalability of transformers on head classes and utilizing inductive biases of CNN for tail classes. To the best of our knowledge, our proposed DeiT-LT for transformers is the \emph{first work in literature that can achieve SotA performance for long-tailed data on small datasets when trained from scratch}. The other works~\cite{LiVT} require transformer pre-training on large datasets, such as ImageNet, to achieve comparable performance on these small datasets.

\vspace{1mm} \noindent \textbf{Results on Large Scale Datasets.}
In this section, we present results attained by DeiT-LT on the large-scale long-tailed datasets of ImageNet-LT and iNaturalist-2018. 
We train all transformer-based methods for similar epochs for a particular dataset, to keep the comparison fair across all baselines (See Suppl. Sec. \ref{suppl:training_config}).  Table \ref{tab:imgnet} presents the result on the ImageNet-LT dataset, where we find that when distilling using LDAM+DRW+SAM (L-D-SAM), our DeiT-LT significantly improves by 2.5\% over the teacher network. 
Notably, it can be seen that our DeiT-LT method, when distilling from PaCo+SAM teacher, achieves a 1.6\% performance gain over the already near SotA teacher network. 
Further, the distillation-based DeiT-LT method achieves a significant gain of 21.6\% and 10.7\% over the baseline transformer training methods, ViT and DeiT-III respectively. This demonstrates that improvement due to distillation scales well with an increase in the size of datasets. 
For iNaturalist-2018, we notice an improvement of close to 3\% over the LDAM+DRW+SAM (L-D-SAM) teacher network and an improvement of 1.7\% over the recent PaCo+SAM teacher.
Additionally, we notice a significant improvement over the data-efficient transformer-based baselines. The data-efficient transformer-based methods struggle while modeling the tail classes, which is supplemented via proposed Distillation loss in DeiT-LT. This enables DeiT-LT to work well across all the classes; the head classes benefit from enhanced learning capacity due to scalable Vision Transformer (ViT), and tail classes are learned well via distillation. Our results are superior for both datasets compared to the CNN-based SotA methods, demonstrating the advantage of DeiT-LT. (Refer Suppl. Sec. \ref{suppl:detailed_results} for detailed results.) %

\begin{table}[t]
    \centering
    \caption{Table showing ablations for various components in DeiT-LT for CIFAR-10 LT and CIFAR-100 LT.}
\begin{tabular}{ccc|cc}
  \toprule
   \textbf{OOD Distill} & \textbf{DRW} & \textbf{SAM} & \textbf{C10 LT}  & \textbf{C100 LT} \\ \midrule
  \xmark & \xmark & \xmark & 70.2 & 31.3 \\
  \cmark & \xmark & \xmark & 84.5 & 48.9 \\
  \cmark & \cmark & \xmark & 87.3 & 54.5\\
  \cmark & \cmark & \cmark & 87.5 & 55.6\\
  \bottomrule
  \end{tabular}
\label{tab:ablation_table}
\vspace{-4mm}
\end{table}

\section{Analysis and Discussion}

\vspace{1mm} \noindent \textbf{Visualizations of Attentions.}
Our training methodology ensures that the \ttt{CLS} and \ttt{DIST} representations diverge while training. While the \ttt{CLS} token is trained against the ground truth, it cannot learn efficient representation for tail classes' images due to ViT's inability to train well on small amounts of data. Distilling from a teacher via out-of-distribution data and introducing re-weighting loss helps the \ttt{DIST} token to learn better representation for the images of minority classes as compared to the \ttt{CLS} token. We further corroborate this by comparing the attention visualization obtained through \emph{Attention Rollout}~\cite{abnar2020quantifying}, for the \ttt{CLS} and \ttt{DIST} token on tail images, for ImageNet-LT dataset (as CIFAR-10 is too small) using DeiT-LT. As can be seen in Fig.~\ref{fig:attention-vis}, the \ttt{CLS} and the \ttt{DIST} token focus on different parts of the image. The \ttt{DIST} token is able to identify the patches of interest (high red intensity) for images of tail classes, while the \ttt{CLS} token fails to do so. The diversity in localized regions demonstrates the complementary information present across the \ttt{CLS} and \ttt{DIST} experts, which is in contrast with DeiT, where both the tokens \ttt{CLS} and \ttt{DIST} are quite similar. We compare visualization with different methods in Suppl. Sec. \ref{suppl:vis_attention}.

\vspace{1mm} \noindent \textbf{Ablation Analysis Across DeiT-LT components.}
We analyze the influence of three key components of our DeiT-LT method, namely OOD distillation, training the Tail Expert classifier with DRW loss, and using SAM teacher for distillation. As can be seen in Table ~\ref{tab:ablation_table}, using OOD distillation brings around 14\% and 18\% improvement over DeiT \cite{touvron2021deit} for CIFAR-10 LT and CIFAR-100 LT, respectively, followed by the other two components, which further improve the accuracy by around 3\% and 6.7\% for CIFAR-10 LT and CIFAR-100 LT, respectively.

\vspace{1mm} \noindent \textbf{Analysis across Transformer Variants.}
In this section, we aim to analyze the performance of DeiT-LT across transformer variants having different capacities. 
\begin{table}[t]
    
    \caption{Analysis across transformer capacity for CIFAR-10 LT  and CIFAR-100 LT for DeiT-LT student($\rho = 100$) with PaCo teacher. }
    \begin{tabular}{c|c|ccc}
        \toprule[1pt]
        \rowcolor{Gray} Model & Overall & Head & Mid & Tail \\
        \midrule 
        \multicolumn{5}{c}{CIFAR-10 LT ($\rho = 100$)} \\
        \midrule 
        DeiT-LT  Tiny (Ti) & 80.8&89.7 &75.1 &79.4 \\
        DeiT-LT  Small (S) & 85.5&92.7 &81.5 &83.7 \\

        DeiT-LT  Base (B) &87.5 &94.5 &84.1 &85.0 \\
        \midrule
        \multicolumn{5}{c}{CIFAR-100 LT ($\rho = 100$)} \\
        \midrule
        DeiT-LT  Tiny (Ti) & 49.3& 66.3& 50.0&27.3 \\
        DeiT-LT  Small (S) & 54.3& 72.6& 54.8&31.1 \\

        DeiT-LT  Base (B) &55.6 &73.1 &56.9 &32.1 \\
\bottomrule      
    \end{tabular}
    \label{tab:cifar10_capacity}
    \vspace{-4mm}
\end{table}
For this, we fix the teacher network and training schedules while varying the network sizes. We experiment with the ViT-Ti, ViT-S, and ViT-B architectures, as introduced in the original ViT work~\cite{dosovitskiyimage}. In Table~\ref{tab:cifar10_capacity}, we observe that the proposed DeiT-LT method scales well with the increased capacity of the Transformer network, and leads to performance improvements.

\label{sec:conclusion}
 \vspace{1mm} 
 \noindent\textbf{Limitations.} One limitation of our 
framework is that the learning for tail classes is done mostly through distillation. Hence, the performance on tail classes remains similar (Table ~\ref{tab:imgnet} and ~\ref{tab:inat}) to that of the CNN classifier. Future works can aim to develop adaptive methods that can shift their focus from CNN to ground truth labels, as the CNN feedback saturates.
\vspace{-1mm}
\section{ Conclusion}
\vspace{-1mm}
In this work, we introduce DeiT-LT, a training scheme to train ViTs from scratch on real-world long-tailed datasets efficiently. We reintroduce the idea of knowledge distillation into ViT students via teacher CNN, as it enables effective learning on the tail classes. This distillation component was found to be redundant and removed from the latest DeiT-III.  Further, in DeiT-LT, we introduce out-of-distribution (OOD) distillation via the teacher, in which we pass strongly augmented images to teachers originally trained via mild augmentations for distillation. The distillation loss is re-weighted to enhance the focus on learning from tail classes. This helps make the classification token an expert on the head classes and the distillation token an expert on the tail classes. To improve generality in minority classes, we induce low-rank features in ViT by distilling from teachers trained from Sharpness Aware Minimization (SAM). The proposed DeiT-LT scheme allows ViTs to be trained from scratch as CNNs and achieve performance competitive to SotA without requiring any pre-training on large-datasets.    

\vspace{1mm} \noindent \textbf{Acknowledgements.} Harsh Rangwani is supported by the PMRF Fellowship. We thank Sumukh for the discussions on the draft.  This work is supported by the SERB-STAR Project (STR/2020/000128) and KIAC Grant.

\renewcommand{\thesection}{S.\arabic{section}}  
\renewcommand{\thetable}{S.\arabic{table}}  
\renewcommand{\thefigure}{S.\arabic{figure}}
\setcounter{figure}{0}   
\setcounter{table}{0}
\clearpage
\setcounter{page}{1}
\appendix

\maketitlesupplementary

\addcontentsline{toc}{section}{} %

\part{} %
\parttoc %

\section{Experimental Details}
\label{suppl:experimental}
\subsection{Datasets}
\label{suppl:datasets}
\textbf{CIFAR-10 LT and CIFAR-100 LT.} We use the imbalanced CIFAR-10 and CIFAR-100 datasets with an exponential decay in sample size across classes. This decay is guided by the Imbalance Ratio ($\rho =\frac{\max_i N_{i}}{\min_j N_j}$). For our experiments on CIFAR-10 LT and CIFAR-100 LT, we show the results on $\rho =100$ and $\rho =50$. CIFAR-10 LT comprises 12,406 training images across 10 classes ($\rho$ = 100). Out of the 10 classes, the first 3 classes are considered \textit{Head} classes with more than 1500 images per class, the following 4 classes are \textit{Mid} (medium) classes with more than 250 images each class, and the last 3 classes account for the \textit{Tail} classes, with each class containing less than 250 images each. Following a similar decay, the 100 classes of CIFAR-100 LT (10,847 training samples with $\rho$ = 100) are also divided into three subcategories: the first 36 classes are considered as the \textit{Head} classes, \textit{Mid} contains the following 35 classes, and the remaining 29 classes are labeled as \textit{Tail} classes. Both CIFAR-10 LT and CIFAR-100 LT datasets are evaluated on held-out sets of 10,000 images each, equally distributed across all classes.

\vspace{1mm} \noindent \textbf{ImageNet-LT.} We use the standard LT dataset created out of ImageNet~\cite{russakovsky2015imagenet}. ImageNet-LT consists of 115,846 training images, with 1280 images in the class with the most images and 5 images in the class with the least images. Out of the 1,000 classes sorted in the descending order of sample frequency, we consider classes with more than 100 samples as \textit{Head} classes, the classes with samples between 20 and 100 to be \textit{Mid} classes and the classes with less than 20 samples as the \textit{Tail} classes as done in \citet{cui2021parametric}.

\vspace{1mm} \noindent \textbf{iNaturalist-2018.} iNaturalist-2018~\cite{van2018inaturalist} is a real-world imbalanced dataset with 437,513 training images. Out of the 8,142 classes sorted in the descending order of sample frequency, we consider classes with more than 100 samples as \textit{Head} classes, the classes with samples between 20 and 100 to be \textit{Mid} classes and the classes with less than 20 samples as the \textit{Tail} classes, similar to ImageNet-LT.

\begin{table*}[t]
    \centering
    
    \caption{Summary of our training procedures used to train DeiT-LT Base (B) from scratch on CIFAR-10 LT, CIFAR-100 LT, ImageNet-LT and iNaturalist-2018. 
    }
    \adjustbox{max width=\textwidth}{
    \begin{tabular}{c|c|c|c|c}
        \toprule[1pt]
        \rowcolor{Gray} Procedure & CIFAR-10 LT & CIFAR-100 LT & ImageNet-LT & iNaturalist-2018 \\
        \midrule
        Epochs   &1200&1200&1400&1000 \\
        Optimizer & AdamW &AdamW&AdamW& AdamW\\
        Effective Batch Size  & 1024 & 1024 &2048&2048\\
        LR  & 5$\times10^{-4}$ & 5$\times10^{-4}$ & 5$\times10^{-4}$ & 5$\times10^{-4}$\\
        LR schedule & cosine & cosine & cosine & cosine \\
        Warmup Epochs   & 5 & 5 & 5 & 5 \\
        DRW starting epoch & 1100 & 1100 & 1200 & 900\\
        \midrule
        Mixup ($\alpha$) & 0.8 & 0.8 &0.8&0.8 \\
        Cutmix ($\alpha$) & 1.0 & 1.0 & 1.0 & 1.0 \\
        Mixup and Cutmix during DRW &$\times$& $\times$ &$ \checkmark $&$ \checkmark $ \\
        Horizontal Flip & $\checkmark$ &$\checkmark$&$ \checkmark $&$ \checkmark $ \\
        Color Jitter & $\checkmark$ & $\checkmark$ &$ \checkmark $& $ \checkmark $\\
        Random Erase & $\checkmark$ & $\checkmark$ &$ \times $& $ \times $\\
       Label smoothing & 0.1 & 0.1 &0.1& 0.1\\
       Solarization & $\times$ & $\times$ & $\checkmark$& $\checkmark$\\ 
       Random Grayscale & $\times$ & $\times$ & $\checkmark$ & $\checkmark$ \\
        Repeated Aug & $ \checkmark $ &$ \checkmark $&$\times$&$\times$ \\
        Auto Aug  & $ \checkmark $ &$ \checkmark $&$ \times $&$ \times $ \\

\bottomrule      
    \end{tabular}}
    \label{tab:hyperparams}

\end{table*}

\subsection{Training Configuration}
\label{suppl:training_config}
In this subsection, we detail the strategies adopted to train DeiT-LT Base (B) model on four benchmark datasets, namely CIFAR-10 LT, CIFAR-100 LT, ImageNet-LT, and iNaturalist-2018. We use the AdamW optimizer to train DeiT-LT from scratch across all the datasets. These runs use a cosine learning rate decay schedule with an initial learning rate of 5$\times10^{-4}$. All the runs use a linear learning rate warm-up schedule for the initial five epochs. Furthermore, we deploy label smoothing with $\varepsilon=0.1$ for all our experiments where the ground truth labels are used to train the \texttt{CLS} expert. Under label smoothing, the true label is assigned a $(1-\varepsilon)$ probability, and the remaining $\varepsilon$ is distributed amongst the other labels. We use hard labels as distillation targets from the teacher network to train the \ttt{DIST} expert classifier via distillation from CNN teacher (Fig.~\ref{fig:overview}). For training the teacher networks with SAM optimizer, we follow the setup mentioned in ~\cite{rangwani_escapingsaddle}

\vspace{1mm} \noindent \textbf{CIFAR-10 LT and CIFAR-100 LT }: We train DeiT-LT for 1200 epochs on imbalanced versions of CIFAR datasets. DRW loss is added to the training of the \ttt{DIST} expert classifier after 1100 epochs. Mixup and Cutmix are used during the initial 1100 epochs of the training. As suggested in~\cite{touvron2021deit}, we use Repeated Augmentation to improve the performance of the DeiT-LT training. The (32$\times$ 32) images of CIFAR datasets are resized to (224 $\times$ 224) before feeding into the transformer architecture. For CIFAR-10 LT and CIFAR-100 LT datasets, ResNet-32 is used as the teacher network. The teacher is trained from scratch on these imbalanced datasets using LDAM+DRW+SAM~\cite{rangwani_escapingsaddle} and contrastive PaCo+SAM (training PaCo~\cite{cui2021parametric} with SAM~\cite{foret2020sharpness} optimizer) frameworks. The input images to the teacher are of size (32$\times$ 32), with the same augmentation used as input images to the teacher network during DeiT-LT training.  

\vspace{1mm} \noindent \textbf{ImageNet-LT and iNaturalist-2018.} DeiT-LT is trained from scratch for 1400 epochs on ImageNet-LT and for 1000 epochs on iNaturalist-2018. DRW loss for distillation head (\ttt{DIST} expert classifier) is initialized from epochs 1200 and 900 for ImageNet-LT and iNaturalist-2018, respectively. Mixup and Cutmix are used throughout the training, including the DRW training phase. More details regarding the training configuration can be found in Table \ref{tab:hyperparams}.

For the ImageNet-LT and iNaturalist-2018 datasets, the ResNet-50 teacher is trained from scratch on the respective datasets using the LDAM+DRW+SAM~\cite{rangwani_escapingsaddle} and contrastive PaCo+SAM (training PaCo~\cite{cui2021parametric} with SAM~\cite{foret2020sharpness} optimizer) methods. The input image size is (224 $\times$ 224) for both the student and teacher network.

\subsection{Additional Baselines}
\label{suppl:additional_baselines}
\begin{figure}[!t]
    \centering
    \caption{ Comparison of training loss for vanilla ViT and ViT+LDAM training on CIFAR-10 LT}
    \includegraphics[width=\columnwidth]{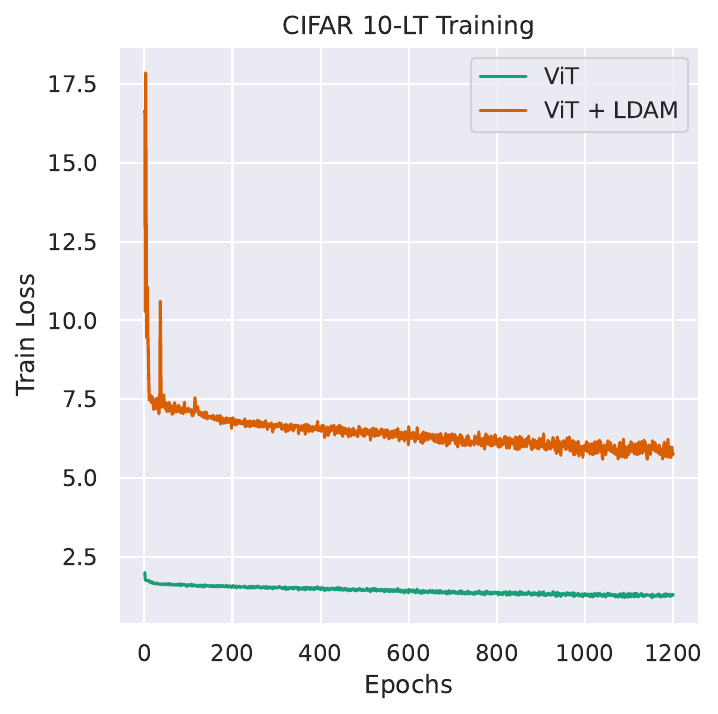}
    
    \label{fig:vit_ldam}
\end{figure}

We want to highlight that we attempted training baselines, like LDAM for vanilla ViT. However, we find that the LDAM baseline (52.75\%) performs inferiorly to the vanilla ViT baselines (62.62\%). We find that the loss for the LDAM baseline gets plateaued very early, and the model does not fit to the training dataset (Fig.~\ref{fig:vit_ldam}). To make the comparison fair with DeiT baselines, we used similar augmentation and other hyperparameters for the ViT Baselines. We think this can be one reason for the non-convergence of the ViT-LDAM baseline. We find that similar abysmal performance for LDAM baseline is also reported by the recent work~\cite{xu2023rethink}, which also resonates with our finding. We think that investigation into this behavior is a good direction for future work.  

 Additionally, for a fair comparison, we do not compare against baselines that use pre-training for long-tailed recognition tasks. RAC~\cite{tian2022vl} uses a ViT-B encoder for their retrieval module with weights obtained from pre-training on ImageNet-21K. The authors do not report on small-scale datasets, as they acknowledge the unfair advantage of using the information present in the pretrained encoder. Similarly, for small-scale datasets, LiVT~\cite{LiVT} method pretrains the encoder via Masked Generative Pretraining on ImageNet-1k. On the contrary, our DeiT-LT method enables training \textit{ViT from scratch} for both small-scale and large-scale datasets.

\subsection{Augmentations for OOD distillation}

\label{suppl:training_aug}

While both DeiT and our DeiT-LT pass images with strong augmentations to the teacher network for distilling into the student network, the set of augmentations used to train the teacher network itself differs between the two approaches. DeiT first trains a large teacher CNN (RegNetY-16GF) using the same set of strong augmentations as that used for the student network. However, we find that distilling from a small teacher CNN (such as ResNet32) trained with weak augmentations gives better performance (see Sec.~\ref{sebsec:out-of-dis-dist})
for more details). Table \ref{tab:teacher_comparison} compares the augmentations used to train the teacher for DeiT (RegNetY-16GF) and for our method DeiT-LT. Our experiments use ResNet32 as the teacher network for CIFAR-10 LT and CIFAR-100 LT, and ResNet50 for the Imagenet-LT and iNaturalist-2018 datasets. For the PaCo teacher, we utilize the mildly strong augmentations used by the PaCo~\cite{cui2021parametric} method itself. We would like to convey, that the PaCo training does not utilize the Mixup and CutMix augmentation in particular while training, which helps us to create OOD samples for this using Mixup and CutMix itself. Distilling via out-of-distribution (OOD) images enables the student to learn the inductive biases of the teacher effectively. This is particularly helpful in improving the performance on the tail classes that have significantly fewer training images. 
\begin{table}[t]
    \centering
    \caption{Comparing augmentation used to train RegNetY-16GF (teacher for DeiT training) and ResNet32 (teacher for DeiT-LT training) for CIFAR-10 LT.}
    \begin{tabular}{c|c|c}
        \toprule[1pt]
         \rowcolor{Gray}& RegNetY-16GF & ResNet32  \\
        \rowcolor{Gray}\multirow{-2}{*}{Procedure}& (Strong) & (Weak) \\
        \midrule
        Image Size & 224$\times$224& 32$\times$32\\
        Random Crop &$\checkmark$&$\checkmark$\\ 
        Horizontal Flip&$\checkmark$ &$\checkmark$ \\
        Mixup ($\alpha$)&0.8&$\times$\\
        Cutmix ($\alpha$) &1.0&$\times$\\
        Color Jitter &0.3&$\times$\\
        Random Erase &$\checkmark$&$\times$\\
        Auto Aug &$\checkmark$&$\times$\\
        Repeated Aug &$\checkmark$&$\times$\\
\bottomrule      
    \end{tabular}
    \label{tab:teacher_comparison}
\end{table}

\section{Detailed Results}
\label{suppl:detailed_results}

\begin{table*}
    \centering
    \caption{Accuracy of expert classifiers on Head, Mid, and Tail classes for CIFAR-10(100) LT.}
    \adjustbox{max width=\textwidth}{
    \begin{tabular}{c|c|c|ccc|c|ccc}
    \toprule[1pt]
         \rowcolor{Gray}& & \multicolumn{4}{c|}{CIFAR-10 LT} & \multicolumn{4}{c}{CIFAR-100 LT} \Tstrut\Bstrut\\
         \cline{3-10}
          \rowcolor{Gray}\multirow{-2}{*}{Imbalance} & \multirow{-2}{*}{Expert} & Overall & Head & Mid & Tail & Overall & Head & Mid & Tail \Tstrut\Bstrut\\
         \midrule[1pt]
         &Average & 87.3$_{\pm0.10}$&93.8$_{\pm0.33}$&83.7$_{\pm0.26}$&85.7$_{\pm0.33}$&54.8$_{\pm0.42}$&72.8$_{\pm0.16}$&55.9$_{\pm0.51}$&31.0$_{\pm0.73}$\Tstrut\Bstrut\\
         \cline{2-10}
         &\texttt{CLS}&78.6$_{\pm0.15}$&96.5$_{\pm0.06}$&79.4$_{\pm0.39}$&59.7$_{\pm0.20}$&43.3$_{\pm0.39}$&73.7$_{\pm0.19}$&41.7$_{\pm0.73}$&7.5$_{\pm0.26}$ \Tstrut\Bstrut\\
         \cline{2-10}
         \multirow{-3}{*}{100}&\texttt{DIST} &79.9$_{\pm0.31}$&72.8$_{\pm0.92}$&75.4$_{\pm0.18}$&93.0$_{\pm0.15}$&42.5$_{\pm0.48}$&39.3$_{\pm1.64}$&45.1$_{\pm0.47}$&43.1$_{\pm0.33}$\Tstrut\Bstrut\\
        \midrule[1pt]
        &Average&89.9$_{\pm0.17}$&94.5$_{\pm0.18}$&87.2$_{\pm0.26}$&88.8$_{\pm0.34}$&60.6$_{\pm0.03}$&74.6$_{\pm0.10}$&60.5$_{\pm0.10}$&43.1$_{\pm0.06}$  \Tstrut\Bstrut\\
         \cline{2-10}
         &\texttt{CLS}&84.1$_{\pm0.33}$&96.5$_{\pm0.12}$&83.3$_{\pm0.66}$&72.8$_{\pm0.55}$&49.6$_{\pm0.21}$&76.0$_{\pm0.31}$&50.5$_{\pm0.46}$&15.9$_{\pm0.41}$  \Tstrut\Bstrut\\
         \cline{2-10}
         \multirow{-3}{*}{50}&\texttt{DIST}&83.2$_{\pm0.23}$&74.6$_{\pm0.51}$&81.8$_{\pm0.21}$&93.6$_{\pm0.08}$&48.0$_{\pm0.20}$&44.0$_{\pm0.25}$&48.4$_{\pm0.36}$&52.6$_{\pm0.07}$  \Tstrut\Bstrut\\

         \bottomrule
         
    \end{tabular}}
    
    \label{tab:expert_performance_CF}
\end{table*}

\begin{table*}[h]
    \centering
    \caption{Accuracy of experts on Head, Mid and Tail classes for ImageNet-LT and iNaturalist-2018.}
    \adjustbox{max width=\textwidth}{
    \begin{tabular}{c|c|ccc|c|ccc}
    \toprule[1pt]
         \rowcolor{Gray}& \multicolumn{4}{c|}{ImageNet-LT} & \multicolumn{4}{c}{iNaturalist-2018} \Tstrut\Bstrut\\
         \cline{2-9}
          \rowcolor{Gray}\multirow{-2}{*}{Expert} & Overall & Head & Mid & Tail & Overall & Head & Mid & Tail \Tstrut\Bstrut\\
          \midrule
         Average & 59.1 & 66.7 & 58.3 & 40.0 & 75.1 & 70.3 & 75.2 & 76.2 \Tstrut\Bstrut\\
         \hline
         \texttt{CLS} expert classifier & 47.5 & 68.3 & 40.0 & 13.5 & 65.6 & 73.8 & 65.8 & 63.1 \Tstrut\Bstrut\\
         \hline
         \texttt{DIST} expert classifier & 56.4 & 57.2 & 58.6 & 46.6 & 72.9 & 56.1 & 73.2 & 77.0 \Tstrut\Bstrut\\
         \bottomrule
    \end{tabular}}
    \label{tab:expert_performance_imnet}
\end{table*}

\textbf{Performance of individual experts}:
Our approach focuses on training diverse experts, where the \ttt{CLS} expert classifier is able to perform well on \textit{Head} (majority) classes, while the \ttt{DIST} expert classifier is able to perform well on the \textit{Tail} (minority) classes. By averaging the output of the individual classifiers, we are able to exploit the benefit of both. 

In this portion, we discuss the individual performance of the \ttt{CLS} and \ttt{DIST} expert classifiers of our proposed DeiT-LT method on CIFAR-10 LT, CIFAR-100 LT, ImageNet-LT, and iNaturalist-2018. As can be seen in Table \ref{tab:expert_performance_CF} and Table \ref{tab:expert_performance_imnet}, the \ttt{CLS} and \ttt{DIST} classifiers give a contrasting performance on the head and tail classes, supporting our claim of expert classifiers. For CIFAR-10 LT ($\rho$ = 100), the \ttt{CLS} expert classifier is able to report an accuracy of 96.5\% on images of the head classes, whereas the \ttt{DIST} expert classifier settles with 72.8\% on the same set of classes. On the other hand, the \ttt{DIST} expert classifier reports 93.0\% accuracy on the tail classes, which is almost 33\% more than that of the \ttt{CLS} expert classifier. Like CIFAR-10 LT, the \ttt{CLS} expert classifier performs better on the head classes of CIFAR-100 LT ($\rho$ = 100) than the \ttt{DIST}, whereas the \ttt{DIST} expert classifier reports much higher accuracy on the tail classes. The \ttt{CLS} classifier achieves an accuracy of 73.7\% on the head classes, and the \ttt{DIST} expert classifier secures 43.1\% accuracy on the tail classes. 
We notice that by averaging the output of the classifiers, we are able to report good performance in both the majority and the minority classes. CIFAR-10 LT reaches an overall accuracy of 87.3\%, with 93.8\% on head classes and 85.7\% on tail classes. Similarly, with 72.8\% on the head and 31.0\% on the tail, DeiT-LT is able to secure an overall 54.8\% on CIFAR-100 LT. The results demonstrate that there is a parallel trend in the performance of experts for both CIFAR-10 LT and CIFAR-100 LT when $\rho$ is set to 50.

A similar trend is seen for large-scale ImageNet-LT and iNaturalist-2018 in Table \ref{tab:expert_performance_imnet}. For ImageNet-LT, the \ttt{CLS} expert classifier reports 68.3\% accuracy on the head classes, which is approximately 11\% more reported by the \ttt{DIST} expert classifier. At the same time, we observe that the \ttt{DIST} expert classifier is able to get an accuracy of 46.6\% on the tail, which is significantly higher than the 13.5\% of the \ttt{CLS} expert classifier. For iNaturalist-2018 as well, the \ttt{CLS} expert classifier achieves a high accuracy of 73.8\% on the head classes, and the \ttt{DIST} expert classifier reaches 77.0\% on the tail classes. After averaging the outputs of the two classifiers, DeiT-LT reports an overall accuracy of 59.1\% for ImageNet-LT and 75.1\% for iNaturalist-2018, which would not have been possible by training a standard Vision Transformer (ViT) with a single classifier.

\section{Comparison with CLIP based methods}
\label{suppl:clip}
Recently, some approaches such as VL-LTR \citep{tian2022vl} and PEL \citep{shi2023parameter} have adopted a pre-trained CLIP backbone to address long-tailed recognition challenges. As indicated originally, and also reinforced by \citep{xu2023demystifying}, CLIP is trained on large-scale balanced dataset (400 M Image-Text pair). As there is a lot of \emph{overlapping concepts between balanced CLIP data and long-tailed datasets (ImageNet-LT and iNat-18)}, the performance of the CLIP fine-tuned methods  \emph{does not indicate meaningful progress on long-tail learning tasks}, as CLIP has already seen tail concepts in abundance. Due to this \underline{unfairness} in training datasets used, we refrain from comparing the CLIP fine-tuned models (i.e., VL-LTR, PEL etc.) with DeiT-LT models trained from scratch.

\section{Visualization of Attention}
\label{suppl:vis_attention}
\begin{figure*}[ht]
    \centering
    \includegraphics[width=\textwidth]{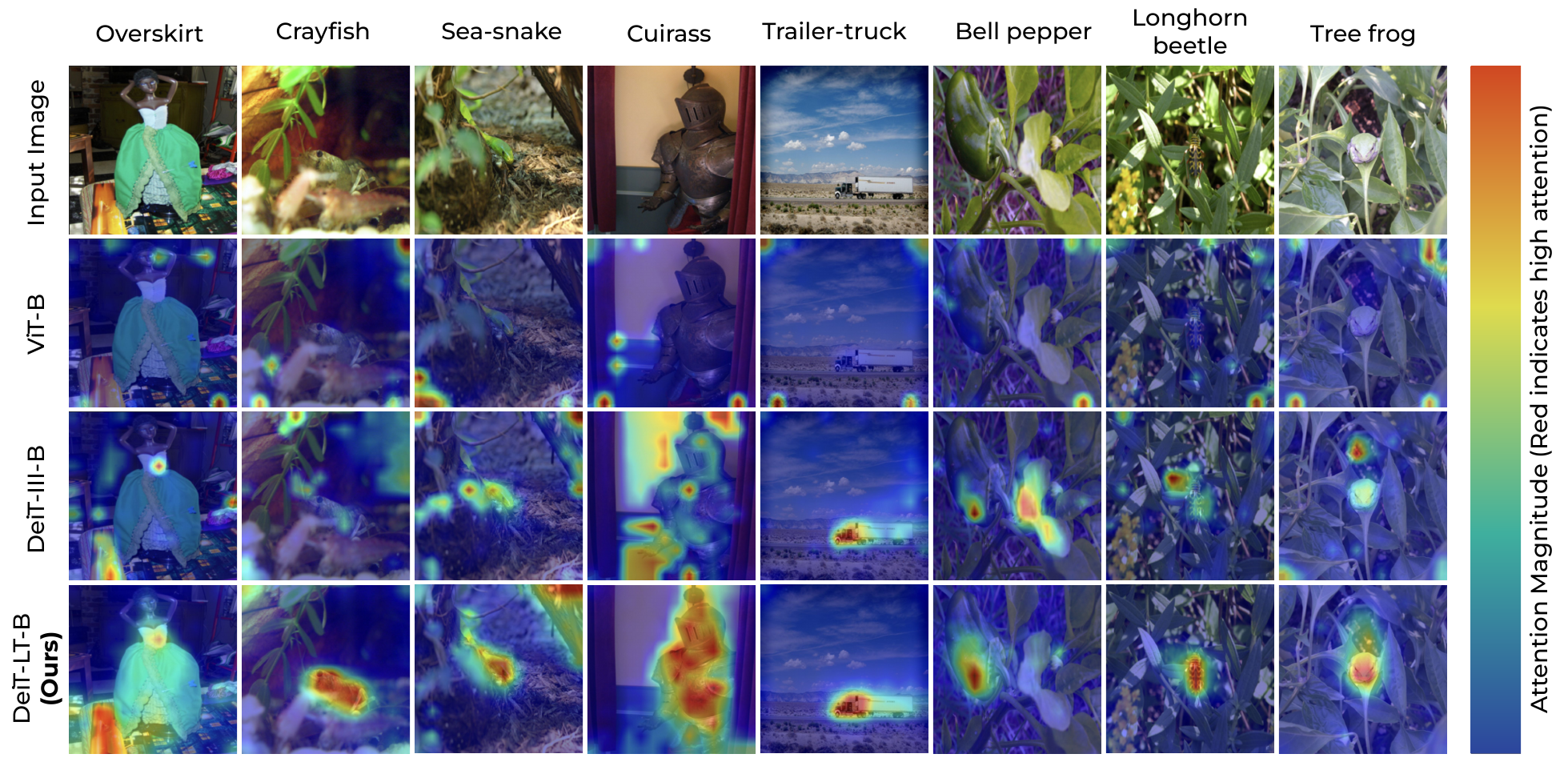}
    \caption{ Visual comparison of the attention maps from ViT-B, DeiT-III \cite{touvron2022deit} and DeiT-LT \emph{(ours)} on the ImageNet-LT dataset,  computed using the method of \emph{Attention Rollout} \cite{abnar2020quantifying}.}
    \label{fig:attention-vis-suppl}
\end{figure*}

To demonstrate the effect of distillation in DeiT-LT, we visualize the attention of baseline methods on ImageNet-LT without distillation (ViT and DeiT-III) and compare it with DeiT-LT. As DeiT-LT contains both the \ttt{DIST} token and the \ttt{CLS} token, for visualization we average the attention across both.  We use the Attention Rollout~\cite{tan2019attention} method for visualization. Fig.~\ref{fig:attention-vis-suppl} shows the result of attention for different methods. It can be clearly observed that DeiT-LT is able to localize attention at the correct position of objects, across almost all cases. We find that DeiT-III attention maps are better in comparison to ViT, but it also often gets confused (eg. Bell Pepper, Sea Snake etc.) compared to DeiT-LT.

\section{Statistical Significance of Experiments}
\label{suppl:signifance}
In this section, we present the results of our experiments on CIFAR-10 LT and CIFAR-100 LT ($\rho$ = 100, 50)(as in Table~\ref{tab:expert_performance_CF}), with three different random seeds. In Table~\ref{tab:expert_performance_CF}, we report the average performance of the expert classifiers along with the standard error for each. The low error demonstrates that the DeiT-LT training procedure is stable and quite robust across random seeds.

\section{Details on Local Connectivity Analysis}
\label{suppl:local_connectivity}
\begin{figure}[!t]
    \centering
    \includegraphics[width=\columnwidth]{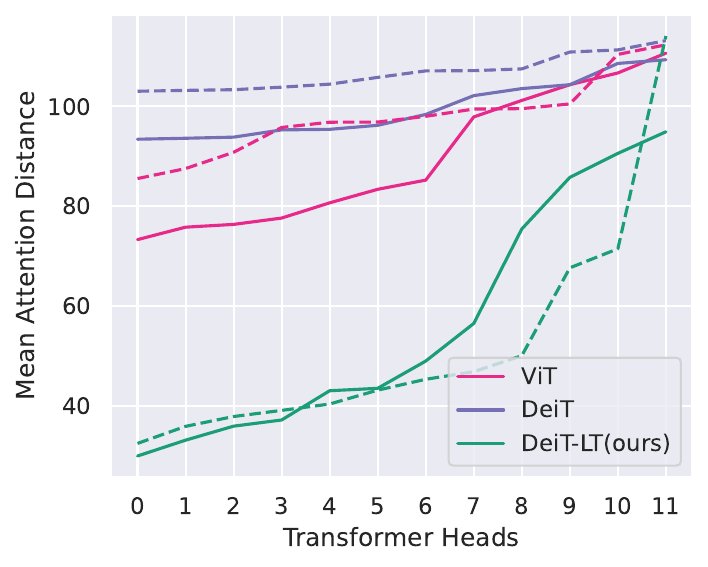}
    \caption{Mean attention distance for early blocks (1,2) for CIFAR-100 LT tail training images.}
    \label{fig:local_cifar-100}
\end{figure}

We compute the mean attention distance for samples of tail classes (i.e. 7,8,9 class for CIFAR-10) using the method proposed by \citet{raghu2021vision}. For each head present in self-attention blocks, we calculate the distance of the patches it attends to. More specifically, we weigh the distance in the pixel space with the attention value and then average it. This is averaged for all the images present in the tail classes. We utilize the code provided here as our reference~\footnote{https://github.com/sayakpaul/probing-vits}. We show in Fig. ~\ref{fig:ViT_locality}
that for early blocks (1 and 2) of ViT, the proposed DeiT-LT method contains local features.
As we go from ViT to distilled DeiT to proposed DeiT-LT, we find that features become more local, which explains the generalizability of DeiT-LT for tail classes. To further confirm our observations, we also provide local connectivity plots for the tail classes of the CIFAR-100 dataset (Fig.~\ref{fig:local_cifar-100}). We observe that DeiT-LT produces highly local features. Further, we find that the DeiT baseline (Table~\ref{tab:cifar10_cifar100}), which is inferior to ViT for CIFAR-100, shows the presence of global features. Hence, the local connectivity correlates well with generalization on tail classes. The correlation of locality of features to generalization has also been observed by ~\cite{raghu2021vision}, who find that using the ImageNet-21k dataset for pre-training leads to more local and generalizable features in comparison to networks pre-trained on ImageNet-1k data.

\begin{figure*}[t]
    \begin{subfigure}{0.49\textwidth}
    \centering
         \centering
         \includegraphics[width=\textwidth]{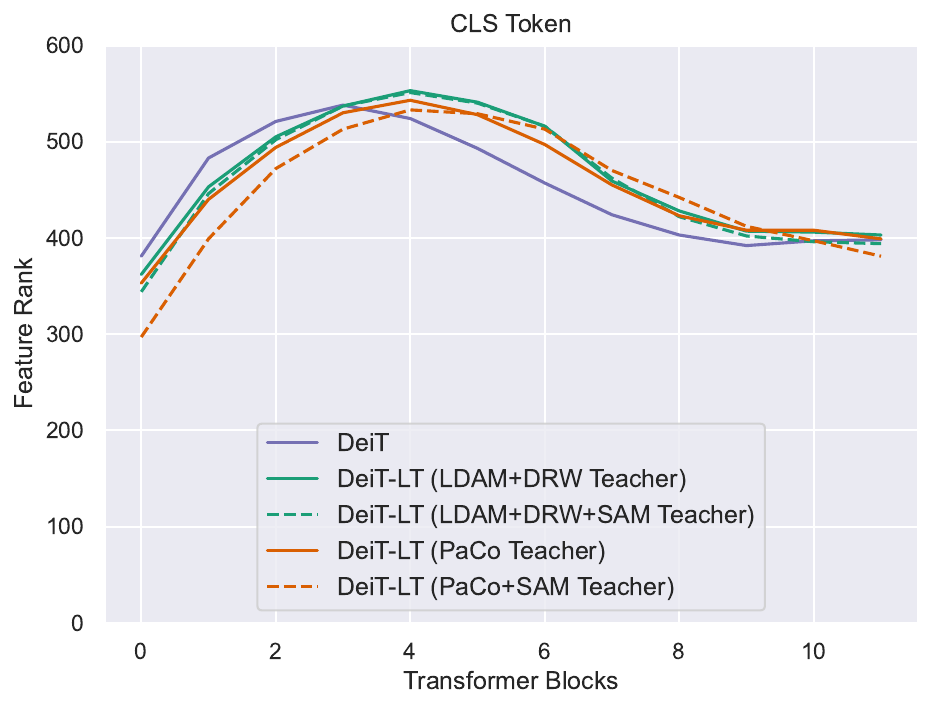}
         \caption{Rank of ViT from Distillation of CNN teachers using \texttt{CLS} token}
         \label{fig:supp_cls_vit_rank}
    
    \end{subfigure}%
   \begin{subfigure}{0.49\textwidth}
         \centering
         \includegraphics[width=\textwidth]{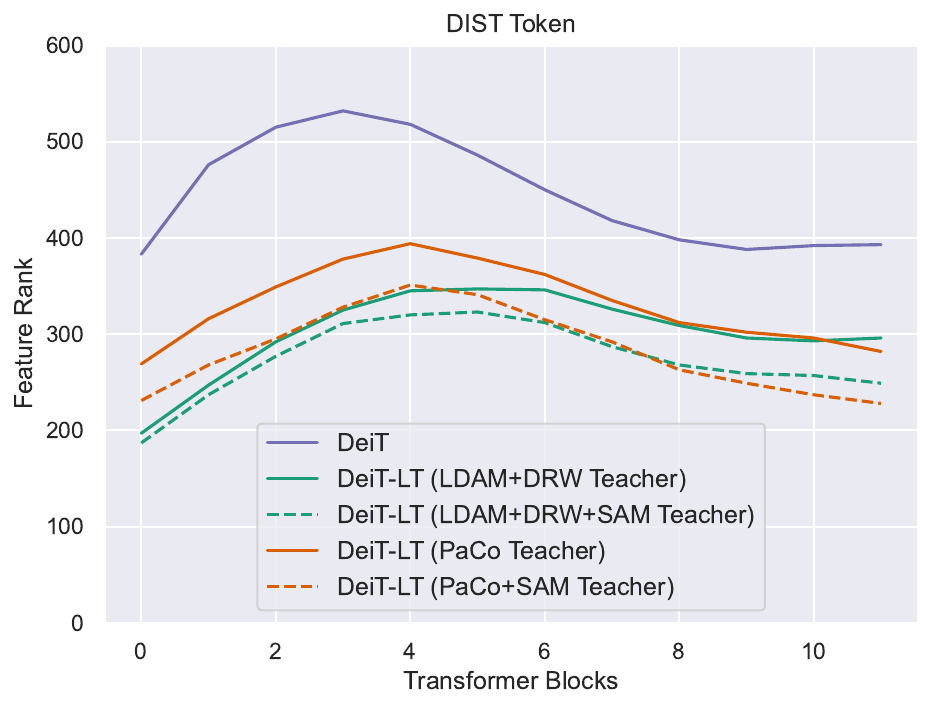}
         \caption{Rank of ViT from Distillation of CNN teachers using \texttt{DIST} token}
         \label{fig:supp_dist_ViT_rank}
    \end{subfigure} 
    \caption{We compare the rank calculated using features from the a) \texttt{CLS} token and b) \texttt{DIST} token when trained on CIFAR-10 LT. Our DeiT-LT captures both fine-grained features (from high-rank \texttt{CLS} token) and generalizable features (from low-rank \texttt{DIST} token).  }
    \label{fig:supp_rank}
\end{figure*}

\begin{figure*}[h]
\vspace{5mm}
\centering

\includegraphics[width=7cm,height=5.5cm]{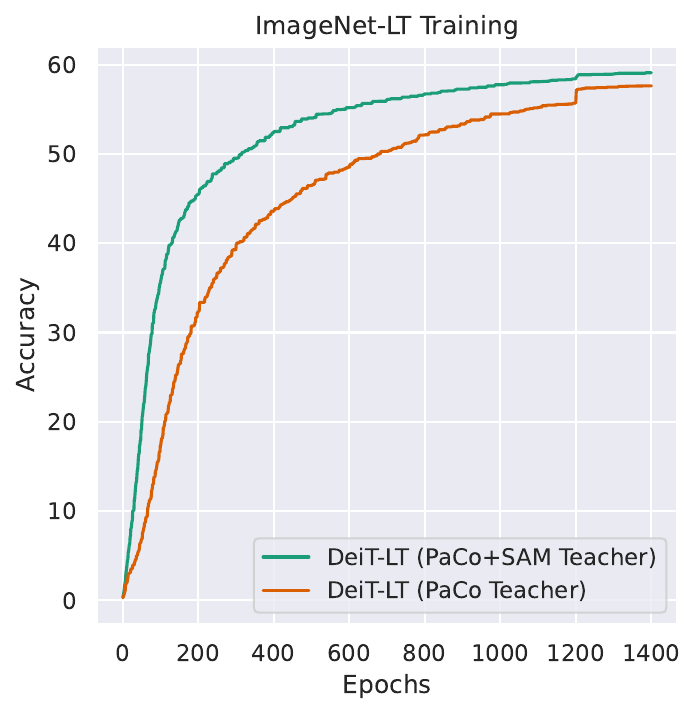}
\label{fig:test2}
\includegraphics[width=7cm,height=5.5cm]{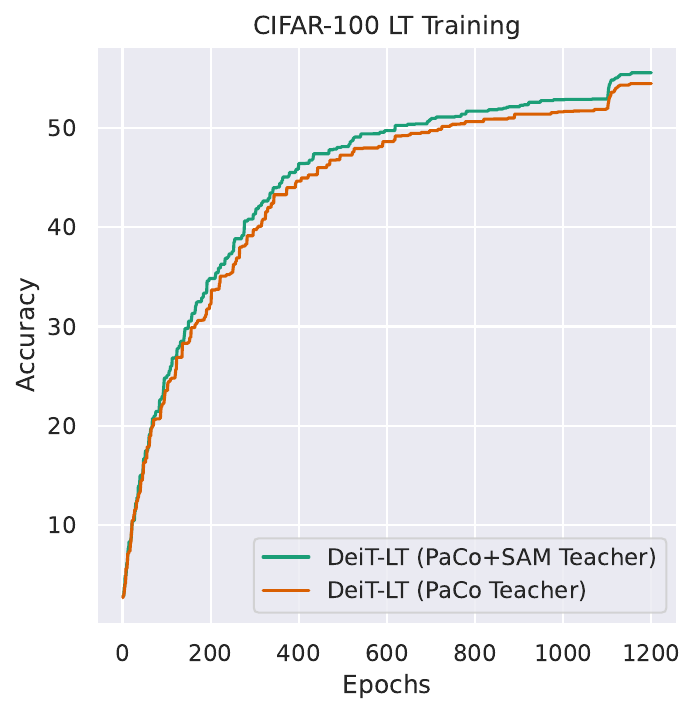}
\caption{Validation Accuracy Plots for the ImageNet-LT \emph{(left)} and CIFAR-100 LT \emph{(right)}. It can be observed that DeiT-LT trained with SAM teachers converges faster than vanilla teachers.}
\vspace{5mm}
\label{fig:convg-plots}
\end{figure*}
\section{Distilling low-rank features}
\label{suppl:low_rank}
In our proposed method, as the \texttt{DIST} token serves as the expert on tail classes, it is important to ensure that it learns generalizable features for minority classes that are less prone to overfitting. As stated in \cite{andriushchenko2023sharpness}, training a network with SAM optimizer leads to low-rank features. In this subsection, we investigate the feature rank of the \texttt{DIST} token that is distilled via a SAM-based teacher.

\vspace{1mm} \noindent \textbf{Calculating Feature Rank.} Consider two sets of images \(\mathcal{X}_{all}, \mathcal{X}_{min} \subset \mathcal{X}\), where \(\mathcal{X}_{all}, \mathcal{X}_{min}\) refer to the set of images from all the classes and minority (tail) classes, respectively, with
\(\mathcal{X}\) being the set of all images.  We construct feature matrices \(F_{n_{h}, d}^{all}\) and \(F_{n_{t}, d}^{min}\), where \(n_h\) and \(n_t\) are the number of images in \(\mathcal{X}_{all}\) and \(\mathcal{X}_{min}\) respectively, and \(d\) is the dimension of the feature representation from \texttt{DIST} token.

Upon centering the columns of \(F_{n_{h}, d}^{all}\), we decompose the  feature matrix as \(U, S, V^T = \mathsf{SVD}(F_{n_{h}, d}^{all})\), and project \(F_{n_{t}, d}^{min}\) using the right singular vectors \(V\) as
\begin{equation*}
F_{proj}^{min}(k) = F_{n_{t}, d}^{min} * V_k
\end{equation*}

where \(V_k\) contains the top \(k\) singular vectors (principal componenets). We calculate our rank as the least \(k\) that satisfies

\begin{equation*}
\frac{{|| F_{n_{t}, d}^{min} - F_{recon}^{min}(k)||}^2 }{{|| F_{n_{t}, d}^{min} ||}^2} 	\le 0.01
\end{equation*}

where \(F_{recon}^{min}(k)\) is an approximate reconstructed feature matrix given by \(F_{recon}^{min} (k)= F_{proj}^{min}(k) * {V_k}^T\). \\

As shown in Fig.~\ref{fig:supp_dist_ViT_rank}, we find that the \texttt{DIST} token trained with a SAM-based teacher reports a lower rank. As we are able to use the same principal components to represent both the majority and minority classes' feature representation, it signifies that the \texttt{DIST} token learns generalizable characteristics relevant across different categories of images in an imbalanced dataset. By learning semantic similar features, our training of \texttt{DIST} token ensures good representation learning for minority classes by leveraging the discriminative features learned from majority classes. 

On the other hand, we observe that \texttt{CLS} token learns high-rank feature representations (Fig.~\ref{fig:supp_cls_vit_rank}), signifying that it captures intricately detailed information. Our DeiT-LT, thus, captures a wide range of information by using the predictions made using both fine-grained details from \texttt{CLS} token and generalizable features from \texttt{DIST} token.

\subsection{Convergence Analysis with SAM Teachers}
\label{suppl:sam_convergence}
We find that models distilled from the teachers trained using SAM~\cite{foret2020sharpness} converge faster than the usual CNN teachers. We provide the analysis for the Deit-LT(PaCo+SAM) and DeiT-LT(PaCo) on the ImageNet-LT and CIFAR-100 datasets in Fig.~\ref{fig:convg-plots}. We observe that models with SAM, coverage much faster, particularly for the ImageNet-LT dataset, demonstrating the increased convergence speed for the distillation. This can be attributed to the fact that low-rank models are simpler in structure and are much easier to distill to the transformer.

\section{Computation Requirement}

For training our proposed DeiT-LT method on CIFAR-10 LT and CIFAR-100 LT, we use two NVIDIA RTX 3090 GPU cards with 24 GiB memory each, with both datasets requiring about 15 hours to train to train the ViT student. We train the DeiT-LT student network on four NVIDIA RTX A5000 GPU cards for the large-scale ImageNet-LT dataset and on four NVIDIA A100 GPU cards for the iNaturalist-2018 dataset, in 61 and 63 hours, respectively. 

\clearpage
{
    \small
    \bibliographystyle{ieeenat_fullname}
    \bibliography{main}

\begin{thebibliography}{68}
\providecommand{\natexlab}[1]{#1}
\providecommand{\url}[1]{\texttt{#1}}
\expandafter\ifx\csname urlstyle\endcsname\relax
  \providecommand{\doi}[1]{doi: #1}\else
  \providecommand{\doi}{doi: \begingroup \urlstyle{rm}\Url}\fi

\bibitem[Abnar and Zuidema(2020)]{abnar2020quantifying}
Samira Abnar and Willem Zuidema.
\newblock Quantifying attention flow in transformers.
\newblock \emph{arXiv preprint arXiv:2005.00928}, 2020.

\bibitem[Agarwal et~al.(2021)Agarwal, Krueger, Clark, Radford, Kim, and Brundage]{agarwal2021evaluating}
Sandhini Agarwal, Gretchen Krueger, Jack Clark, Alec Radford, Jong~Wook Kim, and Miles Brundage.
\newblock Evaluating clip: towards characterization of broader capabilities and downstream implications.
\newblock \emph{arXiv preprint arXiv:2108.02818}, 2021.

\bibitem[Andriushchenko et~al.(2023)Andriushchenko, Bahri, Mobahi, and Flammarion]{andriushchenko2023sharpness}
Maksym Andriushchenko, Dara Bahri, Hossein Mobahi, and Nicolas Flammarion.
\newblock Sharpness-aware minimization leads to low-rank features.
\newblock \emph{arXiv preprint arXiv:2305.16292}, 2023.

\bibitem[Cai et~al.(2021)Cai, Wang, and Hwang]{cai2021ace}
Jiarui Cai, Yizhou Wang, and Jenq-Neng Hwang.
\newblock Ace: Ally complementary experts for solving long-tailed recognition in one-shot.
\newblock In \emph{ICCV}, 2021.

\bibitem[Cao et~al.(2019)Cao, Wei, Gaidon, Arechiga, and Ma]{cao2019learning}
Kaidi Cao, Colin Wei, Adrien Gaidon, Nikos Arechiga, and Tengyu Ma.
\newblock Learning imbalanced datasets with label-distribution-aware margin loss.
\newblock In \emph{NeurIPS}, 2019.

\bibitem[Carion et~al.(2020)Carion, Massa, Synnaeve, Usunier, Kirillov, and Zagoruyko]{carion2020end}
Nicolas Carion, Francisco Massa, Gabriel Synnaeve, Nicolas Usunier, Alexander Kirillov, and Sergey Zagoruyko.
\newblock End-to-end object detection with transformers.
\newblock In \emph{European conference on computer vision}, pages 213--229. Springer, 2020.

\bibitem[Chen et~al.(2022)Chen, Agarwal, Abdelkarim, Zhu, and Elhoseiny]{chen2022reltransformer}
Jun Chen, Aniket Agarwal, Sherif Abdelkarim, Deyao Zhu, and Mohamed Elhoseiny.
\newblock Reltransformer: A transformer-based long-tail visual relationship recognition.
\newblock In \emph{CVPR}, 2022.

\bibitem[Cui et~al.(2021)Cui, Zhong, Liu, Yu, and Jia]{cui2021parametric}
Jiequan Cui, Zhisheng Zhong, Shu Liu, Bei Yu, and Jiaya Jia.
\newblock Parametric contrastive learning.
\newblock In \emph{ICCV}, 2021.

\bibitem[Cui et~al.(2019)Cui, Jia, Lin, Song, and Belongie]{cui2019class}
Yin Cui, Menglin Jia, Tsung-Yi Lin, Yang Song, and Serge Belongie.
\newblock Class-balanced loss based on effective number of samples.
\newblock In \emph{CVPR}, 2019.

\bibitem[Deng et~al.(2009)Deng, Dong, Socher, Li, Li, and Fei-Fei]{deng2009imagenet}
Jia Deng, Wei Dong, Richard Socher, Li-Jia Li, Kai Li, and Li Fei-Fei.
\newblock Imagenet: A large-scale hierarchical image database.
\newblock In \emph{CVPR}, 2009.

\bibitem[Dosovitskiy et~al.(2015)Dosovitskiy, Fischer, Springenberg, Riedmiller, and Brox]{dosovitskiy2015discriminative}
Alexey Dosovitskiy, Philipp Fischer, Jost~Tobias Springenberg, Martin Riedmiller, and Thomas Brox.
\newblock Discriminative unsupervised feature learning with exemplar convolutional neural networks.
\newblock \emph{IEEE TPAMI}, 2015.

\bibitem[Dosovitskiy et~al.(2021)Dosovitskiy, Beyer, Kolesnikov, Weissenborn, Zhai, Unterthiner, Dehghani, Minderer, Heigold, Gelly, et~al.]{dosovitskiyimage}
Alexey Dosovitskiy, Lucas Beyer, Alexander Kolesnikov, Dirk Weissenborn, Xiaohua Zhai, Thomas Unterthiner, Mostafa Dehghani, Matthias Minderer, Georg Heigold, Sylvain Gelly, et~al.
\newblock An image is worth 16x16 words: Transformers for image recognition at scale.
\newblock In \emph{ICLR}, 2021.

\bibitem[Foret et~al.(2020)Foret, Kleiner, Mobahi, and Neyshabur]{foret2020sharpness}
Pierre Foret, Ariel Kleiner, Hossein Mobahi, and Behnam Neyshabur.
\newblock Sharpness-aware minimization for efficiently improving generalization.
\newblock \emph{arXiv preprint arXiv:2010.01412}, 2020.

\bibitem[Gupta et~al.(2019)Gupta, Dollar, and Girshick]{gupta2019lvis}
Agrim Gupta, Piotr Dollar, and Ross Girshick.
\newblock {LVIS}: A dataset for large vocabulary instance segmentation.
\newblock In \emph{Proceedings of the {IEEE} Conference on Computer Vision and Pattern Recognition}, 2019.

\bibitem[He et~al.(2016)He, Zhang, Ren, and Sun]{he2016deep}
Kaiming He, Xiangyu Zhang, Shaoqing Ren, and Jian Sun.
\newblock Deep residual learning for image recognition.
\newblock In \emph{CVPR}, 2016.

\bibitem[Hong et~al.(2021)Hong, Han, Choi, Seo, Kim, and Chang]{hong2021disentangling}
Youngkyu Hong, Seungju Han, Kwanghee Choi, Seokjun Seo, Beomsu Kim, and Buru Chang.
\newblock Disentangling label distribution for long-tailed visual recognition.
\newblock In \emph{CVPR}, 2021.

\bibitem[Hong et~al.(2022)Hong, Zhang, Sun, and Yan]{hong2022safa}
Yan Hong, Jianfu Zhang, Zhongyi Sun, and Ke Yan.
\newblock Safa:sample-adaptive feature augmentation for long-tailed image classification.
\newblock In \emph{ECCV}, 2022.

\bibitem[Iscen et~al.(2021)Iscen, Araujo, Gong, and Schmid]{iscen2021cbd}
Ahmet Iscen, Andr{\'e} Araujo, Boqing Gong, and Cordelia Schmid.
\newblock Class-balanced distillation for long-tailed visual recognition.
\newblock 2021.

\bibitem[Jamal et~al.(2020)Jamal, Brown, Yang, Wang, and Gong]{jamal2020rethinking}
Muhammad~Abdullah Jamal, Matthew Brown, Ming-Hsuan Yang, Liqiang Wang, and Boqing Gong.
\newblock Rethinking class-balanced methods for long-tailed visual recognition from a domain adaptation perspective.
\newblock In \emph{CVPR}, 2020.

\bibitem[Kang et~al.(2019)Kang, Xie, Rohrbach, Yan, Gordo, Feng, and Kalantidis]{kang2019decoupling}
Bingyi Kang, Saining Xie, Marcus Rohrbach, Zhicheng Yan, Albert Gordo, Jiashi Feng, and Yannis Kalantidis.
\newblock Decoupling representation and classifier for long-tailed recognition.
\newblock \emph{arXiv preprint arXiv:1910.09217}, 2019.

\bibitem[Kang et~al.(2020)Kang, Li, Xie, Yuan, and Feng]{kang2020exploring}
Bingyi Kang, Yu Li, Sa Xie, Zehuan Yuan, and Jiashi Feng.
\newblock Exploring balanced feature spaces for representation learning.
\newblock In \emph{International Conference on Learning Representations}, 2020.

\bibitem[Kim et~al.(2020)Kim, Jeong, and Shin]{kim2020m2m}
Jaehyung Kim, Jongheon Jeong, and Jinwoo Shin.
\newblock M2m: Imbalanced classification via major-to-minor translation.
\newblock In \emph{CVPR}, 2020.

\bibitem[Kini et~al.(2021)Kini, Paraskevas, Oymak, and Thrampoulidis]{kinivs}
Ganesh~Ramachandra Kini, Orestis Paraskevas, Samet Oymak, and Christos Thrampoulidis.
\newblock Label-imbalanced and group-sensitive classification under overparameterization.
\newblock In \emph{Advances in Neural Information Processing Systems}, pages 18970--18983. Curran Associates, Inc., 2021.

\bibitem[Krizhevsky et~al.(2009)Krizhevsky, Hinton, et~al.]{krizhevsky2009learning}
Alex Krizhevsky, Geoffrey Hinton, et~al.
\newblock Learning multiple layers of features from tiny images.
\newblock 2009.

\bibitem[Li et~al.(2022{\natexlab{a}})Li, Tan, Wan, Lei, and Guo]{li2022nested}
Jun Li, Zichang Tan, Jun Wan, Zhen Lei, and Guodong Guo.
\newblock Nested collaborative learning for long-tailed visual recognition.
\newblock In \emph{CVPR}, pages 6949--6958, 2022{\natexlab{a}}.

\bibitem[Li et~al.(2022{\natexlab{b}})Li, Cheung, and Lu]{gcl}
Mengke Li, Yiu-ming Cheung, and Yang Lu.
\newblock Long tail visual recognition via gaussian clouded logit adjustment.
\newblock In \emph{CVPR}, 2022{\natexlab{b}}.

\bibitem[Li et~al.(2021)Li, Wang, and Wu]{li2021self_iccv}
Tianhao Li, Limin Wang, and Gangshan Wu.
\newblock Self supervision to distillation for long-tailed visual recognition.
\newblock In \emph{ICCV}, 2021.

\bibitem[Li et~al.(2022{\natexlab{c}})Li, Cao, Yuan, Fan, Yang, Feris, Indyk, and Katabi]{li2022targeted}
Tianhong Li, Peng Cao, Yuan Yuan, Lijie Fan, Yuzhe Yang, Rogerio~S Feris, Piotr Indyk, and Dina Katabi.
\newblock Targeted supervised contrastive learning for long-tailed recognition.
\newblock In \emph{Proceedings of the IEEE/CVF Conference on Computer Vision and Pattern Recognition}, pages 6918--6928, 2022{\natexlab{c}}.

\bibitem[Liu et~al.(2019)Liu, Miao, Zhan, Wang, Gong, and Yu]{liu2019large}
Ziwei Liu, Zhongqi Miao, Xiaohang Zhan, Jiayun Wang, Boqing Gong, and Stella~X Yu.
\newblock Large-scale long-tailed recognition in an open world.
\newblock In \emph{CVPR}, 2019.

\bibitem[Long et~al.(2022)Long, Yin, Ajanthan, Nguyen, Purkait, Garg, Blair, Shen, and van~den Hengel]{long2022retrieval}
Alexander Long, Wei Yin, Thalaiyasingam Ajanthan, Vu Nguyen, Pulak Purkait, Ravi Garg, Alan Blair, Chunhua Shen, and Anton van~den Hengel.
\newblock Retrieval augmented classification for long-tail visual recognition.
\newblock In \emph{CVPR}, 2022.

\bibitem[Loshchilov and Hutter(2017)]{loshchilov2017decoupled}
Ilya Loshchilov and Frank Hutter.
\newblock Decoupled weight decay regularization.
\newblock \emph{arXiv preprint arXiv:1711.05101}, 2017.

\bibitem[Menon et~al.(2020)Menon, Jayasumana, Rawat, Jain, Veit, and Kumar]{menon2020long}
Aditya~Krishna Menon, Sadeep Jayasumana, Ankit~Singh Rawat, Himanshu Jain, Andreas Veit, and Sanjiv Kumar.
\newblock Long-tail learning via logit adjustment.
\newblock \emph{arXiv preprint arXiv:2007.07314}, 2020.

\bibitem[Nayak et~al.(2021)Nayak, Mopuri, and Chakraborty]{nayak2021effectiveness}
Gaurav~Kumar Nayak, Konda~Reddy Mopuri, and Anirban Chakraborty.
\newblock Effectiveness of arbitrary transfer sets for data-free knowledge distillation.
\newblock In \emph{Proceedings of the IEEE/CVF Winter Conference on Applications of Computer Vision}, pages 1430--1438, 2021.

\bibitem[Ousidhoum et~al.(2021)Ousidhoum, Zhao, Fang, Song, and Yeung]{ousidhoum2021probing}
Nedjma Ousidhoum, Xinran Zhao, Tianqing Fang, Yangqiu Song, and Dit-Yan Yeung.
\newblock Probing toxic content in large pre-trained language models.
\newblock In \emph{Proceedings of the 59th Annual Meeting of the Association for Computational Linguistics and the 11th International Joint Conference on Natural Language Processing (Volume 1: Long Papers)}, pages 4262--4274, 2021.

\bibitem[Radosavovic et~al.(2020)Radosavovic, Kosaraju, Girshick, He, and Doll{\'a}r]{radosavovic2020designing}
Ilija Radosavovic, Raj~Prateek Kosaraju, Ross Girshick, Kaiming He, and Piotr Doll{\'a}r.
\newblock Designing network design spaces.
\newblock In \emph{Proceedings of the IEEE/CVF conference on computer vision and pattern recognition}, pages 10428--10436, 2020.

\bibitem[Raghu et~al.(2021)Raghu, Unterthiner, Kornblith, Zhang, and Dosovitskiy]{raghu2021vision}
Maithra Raghu, Thomas Unterthiner, Simon Kornblith, Chiyuan Zhang, and Alexey Dosovitskiy.
\newblock Do vision transformers see like convolutional neural networks?
\newblock \emph{Advances in Neural Information Processing Systems}, 34:\penalty0 12116--12128, 2021.

\bibitem[Rangwani et~al.(2021)Rangwani, Mopuri, and Babu]{rangwani2021class}
Harsh Rangwani, Konda~Reddy Mopuri, and R~Venkatesh Babu.
\newblock Class balancing gan with a classifier in the loop.
\newblock In \emph{Conference on Uncertainty in Artificial Intelligence (\textbf{UAI})}, 2021.

\bibitem[Rangwani et~al.(2022{\natexlab{a}})Rangwani, Aithal, Mishra, and R]{rangwani_escapingsaddle}
Harsh Rangwani, Sumukh~K Aithal, Mayank Mishra, and Venkatesh~Babu R.
\newblock Escaping saddle points for effective generalization on class-imbalanced data.
\newblock In \emph{Advances in Neural Information Processing Systems}, pages 22791--22805. Curran Associates, Inc., 2022{\natexlab{a}}.

\bibitem[Rangwani et~al.(2022{\natexlab{b}})Rangwani, Jaswani, Karmali, Jampani, and Babu]{hrangwani2022gsr}
Harsh Rangwani, Naman Jaswani, Tejan Karmali, Varun Jampani, and R.~Venkatesh Babu.
\newblock Improving gans for long-tailed data through group spectral regularization.
\newblock In \emph{European Conference on Computer Vision (\textbf{ECCV})}, 2022{\natexlab{b}}.

\bibitem[Rangwani$^*$ et~al.(2023)Rangwani$^*$, Bansal$^*$, Sharma, Karmali, Jampani, and Babu]{rangwani2023noisytwins}
Harsh Rangwani$^*$, Lavish Bansal$^*$, Kartik Sharma, Tejan Karmali, Varun Jampani, and R.~Venkatesh Babu.
\newblock Noisytwins: Class-consistent and diverse image generation through style{GAN}s.
\newblock In \emph{Conference on Computer Vision and Pattern Recognition (\textbf{CVPR})}, 2023.

\bibitem[Ren et~al.(2020)Ren, Yu, Sheng, Ma, Zhao, Yi, and Li]{ren2020balanced}
Jiawei Ren, Cunjun Yu, Shunan Sheng, Xiao Ma, Haiyu Zhao, Shuai Yi, and Hongsheng Li.
\newblock Balanced meta-softmax for long-tailed visual recognition.
\newblock \emph{arXiv preprint arXiv:2007.10740}, 2020.

\bibitem[Russakovsky et~al.(2015)Russakovsky, Deng, Su, Krause, Satheesh, Ma, Huang, Karpathy, Khosla, Bernstein, et~al.]{russakovsky2015imagenet}
Olga Russakovsky, Jia Deng, Hao Su, Jonathan Krause, Sanjeev Satheesh, Sean Ma, Zhiheng Huang, Andrej Karpathy, Aditya Khosla, Michael Bernstein, et~al.
\newblock Imagenet large scale visual recognition challenge.
\newblock \emph{International journal of computer vision}, 115\penalty0 (3):\penalty0 211--252, 2015.

\bibitem[Shi et~al.(2023)Shi, Wei, Zhou, Han, Shao, and Li]{shi2023parameter}
Jiang-Xin Shi, Tong Wei, Zhi Zhou, Xin-Yan Han, Jie-Jing Shao, and Yu-Feng Li.
\newblock Parameter-efficient long-tailed recognition.
\newblock \emph{arXiv preprint arXiv:2309.10019}, 2023.

\bibitem[Strudel et~al.(2021)Strudel, Garcia, Laptev, and Schmid]{strudel2021segmenter}
Robin Strudel, Ricardo Garcia, Ivan Laptev, and Cordelia Schmid.
\newblock Segmenter: Transformer for semantic segmentation.
\newblock In \emph{Proceedings of the IEEE/CVF international conference on computer vision}, pages 7262--7272, 2021.

\bibitem[Tan et~al.(2019)Tan, Yang, Wan, Hang, Guo, and Li]{tan2019attention}
Zichang Tan, Yang Yang, Jun Wan, Hanyuan Hang, Guodong Guo, and Stan~Z Li.
\newblock Attention-based pedestrian attribute analysis.
\newblock \emph{TIP}, 28\penalty0 (12):\penalty0 6126--6140, 2019.

\bibitem[Tian et~al.(2022)Tian, Wang, Zhu, Dai, and Qiao]{tian2022vl}
Changyao Tian, Wenhai Wang, Xizhou Zhu, Jifeng Dai, and Yu Qiao.
\newblock Vl-ltr: Learning class-wise visual-linguistic representation for long-tailed visual recognition.
\newblock In \emph{European Conference on Computer Vision}, pages 73--91. Springer, 2022.

\bibitem[Tolstikhin et~al.(2021)Tolstikhin, Houlsby, Kolesnikov, Beyer, Zhai, Unterthiner, Yung, Steiner, Keysers, Uszkoreit, et~al.]{tolstikhin2021mlp}
Ilya~O Tolstikhin, Neil Houlsby, Alexander Kolesnikov, Lucas Beyer, Xiaohua Zhai, Thomas Unterthiner, Jessica Yung, Andreas Steiner, Daniel Keysers, Jakob Uszkoreit, et~al.
\newblock Mlp-mixer: An all-mlp architecture for vision.
\newblock \emph{Advances in neural information processing systems}, 34:\penalty0 24261--24272, 2021.

\bibitem[Touvron et~al.(2021{\natexlab{a}})Touvron, Cord, Douze, Massa, Sablayrolles, and Jegou]{touvron2021deit}
Hugo Touvron, Matthieu Cord, Matthijs Douze, Francisco Massa, Alexandre Sablayrolles, and Herve Jegou.
\newblock Training data-efficient image transformers amp; distillation through attention.
\newblock In \emph{International Conference on Machine Learning}, pages 10347--10357, 2021{\natexlab{a}}.

\bibitem[Touvron et~al.(2021{\natexlab{b}})Touvron, Cord, Sablayrolles, Synnaeve, and J{\'e}gou]{touvron2021going}
Hugo Touvron, Matthieu Cord, Alexandre Sablayrolles, Gabriel Synnaeve, and Herv{\'e} J{\'e}gou.
\newblock Going deeper with image transformers.
\newblock In \emph{Proceedings of the IEEE/CVF International Conference on Computer Vision}, pages 32--42, 2021{\natexlab{b}}.

\bibitem[Touvron et~al.(2022{\natexlab{a}})Touvron, Cord, El-Nouby, Verbeek, and Jegou]{Touvron2022ThreeTE}
Hugo Touvron, Matthieu Cord, Alaaeldin El-Nouby, Jakob Verbeek, and Herve Jegou.
\newblock Three things everyone should know about vision transformers.
\newblock \emph{arXiv preprint arXiv:2203.09795}, 2022{\natexlab{a}}.

\bibitem[Touvron et~al.(2022{\natexlab{b}})Touvron, Cord, and J{\'e}gou]{touvron2022deit}
Hugo Touvron, Matthieu Cord, and Herv{\'e} J{\'e}gou.
\newblock Deit iii: Revenge of the vit.
\newblock In \emph{Computer Vision--ECCV 2022: 17th European Conference, Tel Aviv, Israel, October 23--27, 2022, Proceedings, Part XXIV}, pages 516--533. Springer, 2022{\natexlab{b}}.

\bibitem[Van~Horn et~al.(2018)Van~Horn, Mac~Aodha, Song, Cui, Sun, Shepard, Adam, Perona, and Belongie]{van2018inaturalist}
Grant Van~Horn, Oisin Mac~Aodha, Yang Song, Yin Cui, Chen Sun, Alex Shepard, Hartwig Adam, Pietro Perona, and Serge Belongie.
\newblock The inaturalist species classification and detection dataset.
\newblock In \emph{CVPR}, 2018.

\bibitem[Vaswani et~al.(2017)Vaswani, Shazeer, Parmar, Uszkoreit, Jones, Gomez, Kaiser, and Polosukhin]{vaswani2017attention}
Ashish Vaswani, Noam Shazeer, Niki Parmar, Jakob Uszkoreit, Llion Jones, Aidan~N Gomez, {\L}ukasz Kaiser, and Illia Polosukhin.
\newblock Attention is all you need.
\newblock \emph{Advances in neural information processing systems}, 30, 2017.

\bibitem[Wang and Russakovsky(2023)]{wang2023overwriting}
Angelina Wang and Olga Russakovsky.
\newblock Overwriting pretrained bias with finetuning data.
\newblock In \emph{Proceedings of the IEEE/CVF International Conference on Computer Vision}, pages 3957--3968, 2023.

\bibitem[Wang et~al.(2021{\natexlab{a}})Wang, Han, Wei, Zhang, and Wang]{wang2021contrastive}
Peng Wang, Kai Han, Xiu-Shen Wei, Lei Zhang, and Lei Wang.
\newblock Contrastive learning based hybrid networks for long-tailed image classification.
\newblock In \emph{CVPR}, 2021{\natexlab{a}}.

\bibitem[Wang et~al.(2021{\natexlab{b}})Wang, Lian, Miao, Liu, and Yu]{wang2020long}
Xudong Wang, Long Lian, Zhongqi Miao, Ziwei Liu, and Stella~X Yu.
\newblock Long-tailed recognition by routing diverse distribution-aware experts.
\newblock In \emph{ICLR}, 2021{\natexlab{b}}.

\bibitem[Xu et~al.(2023{\natexlab{a}})Xu, Xie, Tan, Huang, Howes, Sharma, Li, Ghosh, Zettlemoyer, and Feichtenhofer]{xu2023demystifying}
Hu Xu, Saining Xie, Xiaoqing~Ellen Tan, Po-Yao Huang, Russell Howes, Vasu Sharma, Shang-Wen Li, Gargi Ghosh, Luke Zettlemoyer, and Christoph Feichtenhofer.
\newblock Demystifying clip data.
\newblock \emph{arXiv preprint arXiv:2309.16671}, 2023{\natexlab{a}}.

\bibitem[Xu et~al.(2023{\natexlab{b}})Xu, Liu, Yang, Chai, and Yuan]{LiVT}
Zhengzhuo Xu, Ruikang Liu, Shuo Yang, Zenghao Chai, and Chun Yuan.
\newblock Learning imbalanced data with vision transformers.
\newblock In \emph{IEEE Conference on Computer Vision and Pattern Recognition (CVPR)}, 2023{\natexlab{b}}.

\bibitem[Xu et~al.(2023{\natexlab{c}})Xu, Yang, Wang, and Yuan]{xu2023rethink}
Zhengzhuo Xu, Shuo Yang, Xingjun Wang, and Chun Yuan.
\newblock Rethink long-tailed recognition with vision transforms.
\newblock In \emph{ICASSP 2023-2023 IEEE International Conference on Acoustics, Speech and Signal Processing (ICASSP)}, pages 1--5. IEEE, 2023{\natexlab{c}}.

\bibitem[Ye et~al.(2020)Ye, Chen, Zhan, and Chao]{ye2020identifying}
Han-Jia Ye, Hong-You Chen, De-Chuan Zhan, and Wei-Lun Chao.
\newblock Identifying and compensating for feature deviation in imbalanced deep learning.
\newblock \emph{arXiv preprint arXiv:2001.01385}, 2020.

\bibitem[Yun et~al.(2019)Yun, Han, Oh, Chun, Choe, and Yoo]{yun2019cutmix}
Sangdoo Yun, Dongyoon Han, Seong~Joon Oh, Sanghyuk Chun, Junsuk Choe, and Youngjoon Yoo.
\newblock Cutmix: Regularization strategy to train strong classifiers with localizable features.
\newblock In \emph{Proceedings of the IEEE/CVF international conference on computer vision}, pages 6023--6032, 2019.

\bibitem[Zhang et~al.(2018)Zhang, Cisse, Dauphin, and Lopez-Paz]{zhang2018mixup}
Hongyi Zhang, Moustapha Cisse, Yann~N. Dauphin, and David Lopez-Paz.
\newblock mixup: Beyond empirical risk minimization.
\newblock In \emph{ICLR}, 2018.

\bibitem[Zhang et~al.(2021{\natexlab{a}})Zhang, Li, Yan, He, and Sun]{zhang2021distribution}
Songyang Zhang, Zeming Li, Shipeng Yan, Xuming He, and Jian Sun.
\newblock Distribution alignment: A unified framework for long-tail visual recognition.
\newblock In \emph{CVPR}, 2021{\natexlab{a}}.

\bibitem[Zhang et~al.(2021{\natexlab{b}})Zhang, Wei, Zhou, and Wu]{zhang2021bag}
Yongshun Zhang, Xiu-Shen Wei, Boyan Zhou, and Jianxin Wu.
\newblock Bag of tricks for long-tailed visual recognition with deep convolutional neural networks.
\newblock In \emph{AAAI}, 2021{\natexlab{b}}.

\bibitem[Zhong et~al.(2021)Zhong, Cui, Liu, and Jia]{zhong2021improving}
Zhisheng Zhong, Jiequan Cui, Shu Liu, and Jiaya Jia.
\newblock Improving calibration for long-tailed recognition.
\newblock In \emph{CVPR}, 2021.

\bibitem[Zhou et~al.(2017)Zhou, Lapedriza, Khosla, Oliva, and Torralba]{zhou2017places}
Bolei Zhou, Agata Lapedriza, Aditya Khosla, Aude Oliva, and Antonio Torralba.
\newblock Places: A 10 million image database for scene recognition.
\newblock \emph{IEEE TPAMI}, 2017.

\bibitem[Zhou et~al.(2020)Zhou, Cui, Wei, and Chen]{zhou2020bbn}
Boyan Zhou, Quan Cui, Xiu-Shen Wei, and Zhao-Min Chen.
\newblock Bbn: Bilateral-branch network with cumulative learning for long-tailed visual recognition.
\newblock In \emph{CVPR}, 2020.

\bibitem[Zhou et~al.(2023)Zhou, Qu, Xu, and Shen]{zhou2023imbsam}
Yixuan Zhou, Yi Qu, Xing Xu, and Hengtao Shen.
\newblock Imbsam: A closer look at sharpness-aware minimization in class-imbalanced recognition.
\newblock In \emph{Proceedings of the IEEE/CVF International Conference on Computer Vision}, pages 11345--11355, 2023.

\end{thebibliography}
}

\end{document}